\documentclass[10pt,twocolumn,letterpaper]{article}

\pdfoutput=1

\usepackage{cvpr}
\usepackage{times}
\usepackage{epsfig}
\usepackage{graphicx}
\usepackage{amsmath}
\usepackage{amssymb}

\usepackage{amsfonts}
\usepackage{amsthm}
\usepackage{bm}
\usepackage{array}
\usepackage[caption=false, font=footnotesize]{subfig}
\usepackage{dblfloatfix}
\usepackage{multicol}
\usepackage{multirow}
\usepackage{footmisc}
\usepackage{xcolor}
\usepackage{colortbl}
\usepackage{cases}
\usepackage{mathrsfs}
\usepackage{overpic}
\usepackage{booktabs}

\definecolor{Gray}{gray}{0.9}
\newcolumntype{g}{>{\columncolor{Gray}}c}

\usepackage[pagebackref=true,breaklinks=true,letterpaper=true,colorlinks,bookmarks=false]{hyperref}

\cvprfinalcopy
\begin{document}

%%%%%%%%% TITLE
\title{Joint Graph-based Depth Refinement and Normal Estimation}

%%%%%%%%% AUTHORS
\author{Mattia Rossi\textsuperscript{*}, Mireille El Gheche\textsuperscript{*}, Andreas Kuhn\textsuperscript{\dag}, Pascal Frossard\textsuperscript{*} \\[5pt]
\textsuperscript{*}\'{E}cole Polytechnique F\'{e}d\'{e}rale de Lausanne, \textsuperscript{\dag}Sony Europe B.V.}

\maketitle
\thispagestyle{empty}

%%%%%%%%%%%%%%%%%%%%%%%%%%%%%%%%%%%%%%%%%%%%%%%%%%%%%%%%%%%%%%%%%%
%%%%%%%%%%%%%%%%%%%%%%%%%%%%%%%%%%%%%%%%%%%%%%%%%%%%%%%%%%%%%%%%%%
%%%%%%%%%%%%%%%%%%%%%%%%%%%%%%%%%%%%%%%%%%%%%%%%%%%%%%%%%%%%%%%%%%
%%%%%%%%%%%%%%%%%%%%%%%%%%%%%%%%%%%%%%%%%%%%%%%%%%%%%%%%%%%%%%%%%%
% Abstract
%
\begin{abstract}
Depth estimation is an essential component in understanding the 3D geometry of a scene, with numerous applications in urban and indoor settings.
These scenarios are characterized by a prevalence of human made structures, which in most of the cases are either inherently piece-wise planar or can be approximated as such.
With these settings in mind, we devise a novel depth refinement framework that aims at recovering the underlying piece-wise planarity of those inverse depth maps associated to piece-wise planar scenes.
We formulate this task as an optimization problem involving a data fidelity term, which minimizes the distance to the noisy and possibly incomplete input inverse depth map, as well as a regularization, which enforces a piece-wise planar solution.
As for the regularization term, we model the inverse depth map pixels as the nodes of a weighted graph, with the weight of the edge between two pixels capturing the likelihood that they belong to the same plane in the scene.
The proposed regularization fits a plane at each pixel automatically, avoiding any a priori estimation of the scene planes, and enforces that strongly connected pixels are assigned to the same plane.
The resulting optimization problem is solved efficiently with the ADAM solver.
Extensive tests show that our method leads to a significant improvement in depth refinement, both visually and numerically, with respect to state-of-the-art algorithms on the Middlebury, KITTI and ETH3D multi-view datasets.
\end{abstract}

%%%%%%%%%%%%%%%%%%%%%%%%%%%%%%%%%%%%%%%%%%%%%%%%%%%%%%%%%%%%%%%%%%
%%%%%%%%%%%%%%%%%%%%%%%%%%%%%%%%%%%%%%%%%%%%%%%%%%%%%%%%%%%%%%%%%%
%%%%%%%%%%%%%%%%%%%%%%%%%%%%%%%%%%%%%%%%%%%%%%%%%%%%%%%%%%%%%%%%%%
%%%%%%%%%%%%%%%%%%%%%%%%%%%%%%%%%%%%%%%%%%%%%%%%%%%%%%%%%%%%%%%%%%
% Introduction
%
\section{Introduction} \label{sec:introduction}

The accurate recovery of depth information in a scene represents a fundamental step for many applications, ranging from 3D imaging to the enhancement of machine vision systems and autonomous navigation.
Typically, \textit{dense depth estimation} is implemented either using active devices such as \textit{Time-Of-Flight} cameras, or via \textit{dense stereo matching} methods that rely on two \cite{zabih_non_param_1994, hirschmuller_stereo_2008, woodford_global_2009, bleyer_patchmatch_2011} or more \cite{furukawa_accurate_2010, jancosek_multi_2011, galliani_massively_2015, schoenberger_pixelwise_2016, huang_deepmvs_2018, yao_mvsnet_2018} images of the same scene to compute its geometry.
Active methods suffer from noisy measurements, possibly caused by light interference or multiple reflections, therefore they can benefit from a post-processing step to refine the depth map.
Similarly, dense stereo matching methods have a limited performance in untextured areas, where the matching becomes ambiguous, or in the presence of occlusions. Therefore, a stereo matching pipeline typically includes a refinement step to fill the missing depth map areas and remove the noise.

In general, the refinement step is guided by the image associated to the measured or estimated depth map.
The depth refinement literature mostly focuses on enforcing some kind of first order smoothness among the depth map pixels, possibly avoiding smoothing across the edges of the guide image, which may correspond to object boundaries \cite{barron_fast_2016, yao_mvsnet_2018, tosi_leveraging_2019}.
Although depth maps are typically piece-wise smooth, first order smoothness is a very general assumptions, which does not exploit the geometrical simplicity of most 3D scenes.
Based on the observation that most human made environments are characterized by planar surfaces, some authors propose to enforce second order smoothness by computing a set of possible planar surfaces a priori and assigning each depth map pixel to one of them \cite{park_asplanar_2019}.
Unfortunately, this refinement strategy imposes to select a finite number of plane candidates a priori, which may not be optimal in practice and lead to reduced performance.

In this article we propose a depth map refinement framework, which promotes a piece-wise planar arrangement of scenes without any a priori knowledge of the planar surfaces in the scenes themselves.
We cast the depth refinement problem into the optimization of a cost function involving a data fidelity term and a regularization.
The former penalizes those solutions deviating from the input depth map in areas where the depth is considered to be reliable, whereas the latter promotes depth maps corresponding to piece-wise planar surfaces.
In particular, our regularization models the depth map pixels as the nodes of a weighted graph, where the weight of the edge between two pixels captures the likelihood that their corresponding points in the 3D scene belong to the same planar surface.

Our contribution is twofold.
On the one hand, we propose a graph-based regularization for depth refinement that promotes the reconstruction of piece-wise planar scenes explicitly.
Moreover, thanks to its underneath graph, our regularization is flexible enough to handle non fully piece-wise planar scenes as well.
On the other hand, our regularization is defined in order to estimate the normal map of the scene jointly with the refined depth map.

The proposed depth refinement and normal estimation framework is potentially very useful in the context of large scale 3D reconstruction \cite{galliani_massively_2015, schoenberger_pixelwise_2016, huang_deepmvs_2018, yao_mvsnet_2018, xu_multi_2019, kuhn_plane_2019, kuhn_deep_2020}, where the large number of images to be processed requires fast dense stereo matching methods, whose noisy and potentially incomplete depth maps can benefit from a subsequent refinement \cite{huang_deepmvs_2018, yao_mvsnet_2018, xu_multi_2019, kuhn_plane_2019, kuhn_deep_2020}.
It is also relevant in the 3D reconstruction fusion step, when multiple depth maps must be merged into a single point cloud and the estimated normals can be used to filter out possible depth outliers \cite{schoenberger_pixelwise_2016, kuhn_deep_2020}.
We test our framework extensively and show that it is effective in both refining the input depth map and estimating the corresponding normal map.

The article is organized as follows.
Section~\ref{sec:related_work} provides an overview on the depth map refinement literature.
Section~\ref{sec:plane_geometry} motivates the novel regularization term and derives the related geometry.
Section~\ref{sec:problem_formulation} presents our problem formulation and Section~\ref{sec:algorithm} presents our full algorithm.
In Section~\ref{sec:experiments} we carry out extensive experiments to test the effectiveness of the proposed depth refinement and normal estimation approach. Section~\ref{sec:conclusions} concludes the paper.

%%%%%%%%%%%%%%%%%%%%%%%%%%%%%%%%%%%%%%%%%%%%%%%%%%%%%%%%%%%%%%%%%%
%%%%%%%%%%%%%%%%%%%%%%%%%%%%%%%%%%%%%%%%%%%%%%%%%%%%%%%%%%%%%%%%%%
%%%%%%%%%%%%%%%%%%%%%%%%%%%%%%%%%%%%%%%%%%%%%%%%%%%%%%%%%%%%%%%%%%
%%%%%%%%%%%%%%%%%%%%%%%%%%%%%%%%%%%%%%%%%%%%%%%%%%%%%%%%%%%%%%%%%%
% Related work
%
\section{Related work} \label{sec:related_work}

Depth refinement methods fall mainly into three classes: local methods, global methods and learning-based methods.

Local methods are characterized by a greedy approach.
Tosi et al.\cite{tosi_leveraging_2019} adopt a two step strategy.
First, the input disparity map is used to compute a binary confidence mask that classifies each pixel as reliable or not.
Then, the disparity at the pixels classified as reliable is kept unchanged and used to infer the disparity at the non reliable ones, using a wise interpolation heuristic.
In particular, for each non reliable pixel, a set of anchor pixels with a reliable disparity is selected and the pixel disparity is estimated as a weighted average of the anchor disparities.
Besides its low computational requirements, the method in \cite{tosi_leveraging_2019} suffers two major drawbacks.
On the one hand, pixels classified as reliable are left unchanged: this does not permit to correct possible pixels misclassified as reliable, which may bias the refinement of the other pixels.
On the other hand, the method in \cite{tosi_leveraging_2019}, and local methods in general, cannot take advantage of the reliable parts of the disparity map fully, due to their local perspective.

Global methods rely on an optimization procedure to refine each pixel of the input disparity map jointly.
Barron and Poole \cite{barron_fast_2016} propose the \textit{Fast Bilateral Solver}, a framework that permits to cast arbitrary image related ill posed problems into a global optimization formulation, whose prior resembles the popular bilateral filter \cite{tomasi_bilateral_1998}.
In \cite{tosi_leveraging_2019} the Fast Bilateral Solver has been shown to be effective in the disparity refinement task, but its general purposefulness prevents it from competing with specialized methods, even local ones like \cite{tosi_leveraging_2019}. Global is also the disparity refinement framework proposed by Park et al. \cite{park_asplanar_2019}, which can be broken down into four steps.
First, the input reference image is partitioned into super-pixels and a \textit{local} plane is estimated for each one of them using RANSAC.
Second, super-pixels are progressively merged into macro super-pixels to cover larger areas of the scene and a new \textit{global} plane is estimated for each of them.
Then, a \textit{Markov Random Field (MRF)} is defined over the set of super-pixels and each one is assigned to one of four different classes: the class associated the local plane of the super-pixel, the class associated to the global plane of the macro super-pixel to which the super-pixel belongs, the class of pixels not belonging to any planar surface, or the class of outliers.
The MRF employs a prior that enforces connected super-pixel to belong to the same class, thus promoting a global consistency of the disparity map.
Finally, the parameters of the plane associated to each super-pixel are slightly perturbed, again within a MRF model, to allow for a finer disparity refinement.
This method is the closest to ours in flavour.
However, the a priori detection of a finite number of candidate planes for the whole scene biases the refinements toward a set of plane hypotheses that may either not be correct, as estimated on the input noisy and possibly incomplete disparity map, or not be rich enough to cover the full geometry of the scene.

Finally, recent learning based methods typically rely on a deep neural network which, fed with the noisy or incomplete disparity map, outputs a refined version of it \cite{gidaris_detect_2016, knobelreiter_learned_2019}.
In \cite{gidaris_detect_2016} the task is split into three sub-tasks, each one addressed by a different network and finally trained end to end as a single one: detection of the non reliable pixels, gross refinement of the disparity map and fine refinement.
Instead, Kn\"{o}belreiter and Pock \cite{knobelreiter_learned_2019} revisit the work of Cherabier at al. \cite{cherabier_learning_2018} in the context of disparity refinement.
First, the disparity refinement task is cast into the minimization of a cost function, hence a global optimization, whose minimizer is the desired refined disparity map.
However, the cost function is partially parametrized, rather than fully handcrafted.
Then, the cost function solver can be unrolled for a fixed number of iterations, thus obtaining a network structure, and the parametrized cost function can be learned.
Once the network parameters are learned, the disparity refinement requires just a network forward pass.
Both the methods in \cite{gidaris_detect_2016} and \cite{knobelreiter_learned_2019} permit a fast refinement of the input disparity.
However, due to their learning-based nature, they can fall short easily in those scenarios which differ from the ones employed at training time, as shown for the method in \cite{knobelreiter_learned_2019}, which performs remarkably well in the Middlebury benchmark \cite{scharstein_high_2014} training set, while quite poorly in the test set of the same dataset.

Our graph-based depth refinement framework, instead, does not rely on any training procedure.
It adopts a global approach which permits to compute a pair of consistent depth and normal maps jointly.
Moreover, it does not need any a priori knowledge of the possible planar surfaces in the scene, as it automatically assigns a plane to each pixel based on its neighbors in the underneath graph.
Finally, the proposed framework does not call for a separate handling of pixels belonging to planar surfaces and not, again thanks to the graph underneath.

%%%%%%%%%%%%%%%%%%%%%%%%%%%%%%%%%%%%%%%%%%%%%%%%%%%%%%%%%%%%%%%%%%
%%%%%%%%%%%%%%%%%%%%%%%%%%%%%%%%%%%%%%%%%%%%%%%%%%%%%%%%%%%%%%%%%%
%%%%%%%%%%%%%%%%%%%%%%%%%%%%%%%%%%%%%%%%%%%%%%%%%%%%%%%%%%%%%%%%%%
%%%%%%%%%%%%%%%%%%%%%%%%%%%%%%%%%%%%%%%%%%%%%%%%%%%%%%%%%%%%%%%%%%
% 3D plane projection within the pinhole camera model
%
\section{Depth map model} \label{sec:plane_geometry}

In this section we investigate the relationship between a plane in the 3D space and its 2D depth map.
In particular, we show that a plane imaged by a camera has an \textit{inverse} depth map described by a plane as well, thus motivating a piece-wise planar model for the inverse depth map of those scenes where planar structures are prevalent.
Let us consider a plane $\mathscr{P}$ in the 3D scene in front of a pinhole camera.
For the sake of simplicity and w.l.o.g., hereafter we assume the scene coordinate system to coincide with the camera one.
In particular, the camera coordinate system is assumed to be left handed, with the $Z$ axis aligned with the camera optical axis and pointing outside the camera.
The plane $\mathscr{P}$ can be described uniquely by a pair $(P_{0}, \bm{n}_{0})$, where $P_{0} = (X_{0}, Y_{0}, Z_{0}) \in \mathscr{P} \subset \mathbb{R}^{3}$ is a point of the plane and $n_{0} = (a_{0}, b_{0}, c_{0}) \in \mathbb{R}^{3}$, with $||n_{0}||_{2} = 1$, is the \textit{plane normal} vector defining the orientation of the plane itself.
Therefore, for all and only the points $P = (X, Y, Z) \in \mathscr{P}$, the following equation holds true:
\begin{equation}\label{eq:plane_equation}
    \langle n_{0}, \left( X, Y, Z \right) - \left( X_{0}, Y_{0}, Z_{0} \right) \rangle = 0,
\end{equation}
where $\langle \cdot, \cdot \rangle$ denotes the dot product.
Equivalently, Eq.~\eqref{eq:plane_equation} can be rewritten as follows:
\begin{equation} \label{eq:plane_equation_2}
    a_{0} X + b_{0} Y + c_{0} Z - \rho_{0} = 0,
\end{equation}
where $\rho_{0} = \langle n_{0}, (X_{0}, Y_{0}, Z_{0}) \rangle \in \mathbb{R}$.
In the pinhole model, the 3D point $P = (X, Y, Z)$ is projected into the camera image plane at the pixel image coordinates $(x, y) \in \mathbb{R}^{2}$:
\begin{equation} \label{eq:pinhole_model}
    x = \frac{X}{Z} f^{x} + c^{x}, \quad y = \frac{Y}{Z} f^{y} + c^{y},
\end{equation}
where $(c^{x}, c^{y}) \in \mathbb{R}^{2}$ are the coordinates of the camera principal point and $f^{x}$, $f^{y}$ are the horizontal and vertical focal lengths, respectively.
Solving for $X$ and $Y$ in Eq.~\eqref{eq:pinhole_model}, the plane equation in Eq.~\eqref{eq:plane_equation_2} can be expressed as a function of the image coordinates ${(x, y)}$ and the corresponding depth ${Z = Z(x, y)}$:
\begin{equation} \label{eq:plane_equation_3}
    \left(
    a_{0} \frac{\left( x - c^{x} \right)}{f^{x}} +
    b_{0} \frac{\left( y - c^{y} \right)}{f^{y}} +
    c_{0} \right) Z - \rho_{0} = 0.
\end{equation}
Similarly, in image coordinates, $\rho_{0}$ reads as follows:
\begin{equation*} \label{eq:rho}
    \rho_{0} = \left(
    a_{0} \frac{\left( x_{0} - c^{x} \right)}{f^{x}} +
    b_{0} \frac{\left( y_{0} - c^{y} \right)}{f^{y}} +
    c_{0} \right) Z_{0},
\end{equation*}
where $(x_{0}, y_{0})$ is the projection of the point $P_{0}$ into the camera image plane.

Let us introduce the vector ${u}(x_{0}, y_{0}) = (u_{0}^{x}, u_{0}^{y}) \in \mathbb{R}^{2}$:
\begin{equation} \label{eq:gradient}
    u_{0}^{x} = \frac{a_{0}}{\rho_{0} f^{x}}, \quad u_{0}^{y} = \frac{b_{0}}{\rho_{0} f^{y}}.
\end{equation}
Using the vector ${u}(x_{0}, y_{0})$ and introducing the \textit{inverse depth} ${d(x, y) = 1 / Z(x, y)}$ permits to rewrite Eq.~\eqref{eq:plane_equation_3} as follows:
\begin{equation} \label{eq:plane_equation_standard}
    d \left( x, y \right) = d \left( x_{0}, y_{0} \right) + \langle u \left( x_{0}, y_{0} \right), \left( x - x_{0}, y - y_{0} \right) \rangle.
\end{equation}
A proof is provided in the supplementary material.
Eq.~\eqref{eq:plane_equation_standard} can be interpreted as a first order Taylor expansion of the inverse depth map at the image coordinate $(x_{0}, y_{0})$, such that $u(x_{0}, y_{0})$ should be close to $\nabla {d}(x_{0}, y_{0})$.
In particular, Eq.~\eqref{eq:plane_equation_standard} shows that the inverse depth $d(x, y)$ of every point $P\in\mathscr{P}$ is described by a plane, which passes through the point $(x_{0}, y_{0}, d(x_{0}, y_{0}))$ and has a plane normal vector $(u(x_{0}, y_{0}), -1) \in \mathbb{R}^{3}$.

We showed that the plane $\mathscr{P}$ is represented equivalently by the pair $(P_{0}, n_{0})$ in the scene domain or by the pair $((x_{0}, y_{0}, d(x_{0}, y_{0})), u(x_{0}, y_{0}))$ in the camera domain.
Eq.~\eqref{eq:gradient} permits to move from the scene domain to the camera one by computing $u(x_{0}, y_{0})$ when $n_{0}$ is given.
To recover $n_{0}$ when $u(x_{0}, y_{0})$ is given instead, we can solve the following non linear system in the variables $a_{0}$, $b_{0}$, $c_{0}$:
\begin{subnumcases}{\label{eq:normal_sys}}
    u_{0}^{x} = \left( \rho_{0} f^{x} \right)^{-1} a_{0} \label{eq:normal_sys_1} \\
    u_{0}^{y} = \left( \rho_{0} f^{y} \right)^{-1} b_{0} \label{eq:normal_sys_2} \\
    a_{0}^{2} + b_{0}^{2} + c_{0}^{2} = 1, \label{eq:normal_sys_3}
\end{subnumcases}
where the constraints in Eqs.~\eqref{eq:normal_sys_1}--\eqref{eq:normal_sys_2} refer to Eq.~\eqref{eq:gradient} and Eq.~\eqref{eq:normal_sys_3} is the normal unitary constraint $||n_{0}||_{2} = 1$.
The closed form solution of the system is provided in the supplementary material.
In what follows, we present our optimization problem to jointly estimate the normal map $u$ and the refined inverse depth map $d$.

%%%%%%%%%%%%%%%%%%%%%%%%%%%%%%%%%%%%%%%%%%%%%%%%%%%%%%%%%%%%%%%%%%
%%%%%%%%%%%%%%%%%%%%%%%%%%%%%%%%%%%%%%%%%%%%%%%%%%%%%%%%%%%%%%%%%%
%%%%%%%%%%%%%%%%%%%%%%%%%%%%%%%%%%%%%%%%%%%%%%%%%%%%%%%%%%%%%%%%%%
%%%%%%%%%%%%%%%%%%%%%%%%%%%%%%%%%%%%%%%%%%%%%%%%%%%%%%%%%%%%%%%%%%
% Problem formulation
%
\section{Depth map refinement problem} \label{sec:problem_formulation}

Given an image ${I}\in\mathbb{R}^{H \times W}$ we are interested in recovering the corresponding depth map ${Z}$ when only a noisy and possibly incomplete estimate $\bar{Z}$ is available.
We assume that $\bar{Z}$ is provided together with a confidence mask $m$ with entries in $[0, 1]$.
In particular, the confidence map is such that, $\forall i = (x, y) \in \{0, \ldots, W-1 \} \times \{0, \ldots, H-1 \}$, $m_{i} = 0$ when the entry $\bar{Z}_{i}$ is considered completely inaccurate, while $m_{i} = 1$ when $\bar{Z}_{i}$ is considered highly accurate.\footnote{A wide variety of algorithms addressing pixel-wise confidence prediction exist in the literature, either based on hand-crafted features or learning-based \cite{poggi_quantitative_2017}. In practice, also the simple stereo reprojection error could be adopted \cite{knobelreiter_learned_2019}.}
In the following, we focus on estimating the refined inverse depth map $d = 1 / Z$ and the corresponding normal map, given the initial estimate $\bar{d} = 1 / \bar{Z}$ and the mask $m$.
We consider the following optimization problem:
\begin{equation} \label{eq:problem_formulation}
    d^{*}, u^{*} \> = \> \underset{d, u}{\mathrm{argmin}} \;\;
    f \left( d \right) + 
    \lambda\, g \left( d, u \right)
\end{equation}
where $f(\cdot)$ is a data term, $g(\cdot)$ is a regularization term for piece-wise planar functions and $\lambda \in \mathbb{R}_{\geq 0}$ is a scalar weight factor.
The refined depth map is eventually computed as $Z^{*} = 1 / d^{*}$, while the 3D normal map $n^{*}$ is obtained from $u^{*}$ via the close form solution of the system in Eq.~\eqref{eq:normal_sys}.

In more details, the data fidelity term $f(\cdot)$ enforces consistency between the inverse depth map estimate $d$ and the input inverse depth map $\bar{d}$.
We adopt a data term of the following form:
\begin{equation*} \label{eq:data_fidelity_d}
    f \left( d \right) =
    \sum_{i} | d_{i} - \bar{d}_{i} | m_{i},
\end{equation*}
which enforces that the estimated inverse depth map $d$ is close to $\bar{d}$ at those pixels $i$ where the latter is considered accurate, i.e., where $m_{i}$ tends to one.\footnote{The quality of the confidence map $m$ can affect the quality of the refined depth map. However, in the case of missing confidence, i.e., $m$ constant, our formulation in Eq.~\eqref{eq:problem_formulation} still promotes piece-wise planar scenes.}

Then, the regularization term $g(\cdot)$ enforces the inverse depth map $d$ to be piece-wise planar, according to the model developed in Section~\ref{sec:plane_geometry}.
In particular, we choose to model the inverse depth map as a weighted graph, where each pixel represents a node and where the weight of the edge between two pixels can be interpreted as the likelihood that the corresponding two points in the 3D scene belong to the same plane.
Namely, if the image looks locally similar at two different pixels, the probability is large for these pixels to belong to the same physical object, hence the same plane.
The regularization term parametrizes the inverse depth at each pixel with a different plane, but it enforces strongly connected pixels in the graph, i.e., those pixels connected by an edge with high weight, to share the same plane parametrization.
Specifically, our regularization term $g(\cdot)$ encompasses two terms balanced by a scalar weight $\alpha \in \mathbb{R}_{>0}$ and reads as follows:
\begin{subequations}
    \begin{align}
        g \left( d, u \right) &=
            \sum_{i} \sqrt{
            \sum_{j \sim i}
            w_{ij}^{2} \left( d_{j} - d_{i} - \langle u_{i}, j - i \rangle \right)^{2}}
            \label{eq:prior_planes} \\
        &+ 
        \alpha \sum_{i} \sum_{j \sim i}
            w_{ij} \Vert u_{j} - u_{i} \Vert_{2},
            \label{eq:prior_normals}
    \end{align}
\end{subequations}
where $\{ j \sim i \}$ describes the direct neighbours of $i$ in the graph and $w_{ij} \in \mathbb{R}_{>0}$ is the weight associated to the edge between the pixel $i$ and its neighbour pixel $j$.

The first term of the regularization in Eq.~\eqref{eq:prior_planes} enforces the following constraint between the pixel $i$ and its neighboring pixel $j$, for every $j \sim i$:
\begin{equation*} \label{eq:prior_planes_explained}
    d_{j} = d_{i}  +  \langle u_{i}, j - i \rangle,
\end{equation*}
which requires the inverse depth map in the neighbourhood of the pixel $i$ to be approximated by the plane $\mathscr{P}_i$ whose orientation is given by the vector $u_{i} \in \mathbb{R}^{2}$.
This constraint recalls Eq.~\eqref{eq:plane_equation_standard} and it is weighted by the likelihood $w_{ij}$ that $i$ and $j$ are the projections of two points $P_{i}$ and $P_{j} \in \mathbb{R}^{3}$ belonging to the same plane $\mathscr{P}_i$ in the 3D scene.
However, using only Eq.~\eqref{eq:prior_planes} does not guarantee that the plane $\mathscr{P}_j$, fitted at the pixel $j$, and the plane $\mathscr{P}_i$ are the same (e.g., the plane normal vectors $u_j$ and $u_i$ coincide).
Therefore, the second term of the regularization in Eq.~\eqref{eq:prior_normals} enforces that the two planes fitted at $i$ and $j$, with orientations $u_{i}$ and $u_{j}$, respectively, are consistent with each other when $P_{i}$ and $P_{j}$ are considered likely to belong to the same plane $\mathscr{P}$.

We conclude by observing that Eq.~\eqref{eq:prior_normals} can be interpreted as a generalization of the well-known anisotropic \textit{Total Variation (TV)} regularization \cite{condat_discrete_2017}, typically referred to as \textit{Non Local Total Variation (NLTV)} \cite{gilboa_nonlocal_2009} in general graph settings.
In fact, the quantity $\| u_{j} - u_{i} \|_{2}$ can be interpreted as the magnitude of the derivative of $u$ at the node $i$ in the direction of the node $j$ \cite{shuman_emerging_2013}, so that Eq.~\eqref{eq:prior_normals} enforces a piece-wise constant signal $u$ on the graph, which enforces the signal $d$ to be piece-wise planar.
This corresponds to the depth map model of Section~\ref{sec:plane_geometry}.

%%%%%%%%%%%%%%%%%%%%%%%%%%%%%%%%%%%%%%%%%%%%%%%%%%%%%%%%%%%%%%%%%%
%%%%%%%%%%%%%%%%%%%%%%%%%%%%%%%%%%%%%%%%%%%%%%%%%%%%%%%%%%%%%%%%%%
%%%%%%%%%%%%%%%%%%%%%%%%%%%%%%%%%%%%%%%%%%%%%%%%%%%%%%%%%%%%%%%%%%
%%%%%%%%%%%%%%%%%%%%%%%%%%%%%%%%%%%%%%%%%%%%%%%%%%%%%%%%%%%%%%%%%%
% Depth refinement algorithm
%
\section{Depth refinement algorithm} \label{sec:algorithm}

In this section we present the structure of the graph underneath the regularization in Eqs.~\eqref{eq:prior_planes}--\eqref{eq:prior_normals}.
Then, we detail the optimization algorithm adopted to find the solution of the joint depth refinement and normal estimation problem presented in Eq.~\eqref{eq:problem_formulation}.

%%%%%%%%%%%%%%%%%%%%%%%%%%%%%%%%%%%%%%%%%%%%%%%%%%%%%%%%%%%%%%%%%%%%%%%%%%%%%%%%%%%%%%%%%%%%%%%%%%%%
%%%%%%%%%%%%%%%%%%%%%%%%%%%%%%%%%%%%%%%%%%%%%%%%%%%%%%%%%%%%%%%%%%%%%%%%%%%%%%%%%%%%%%%%%%%%%%%%%%%%
\subsection{Graph construction} \label{subsec:graph}

We assume that areas of the image ${I}$ sharing the same texture correspond to the same object and likewise to the same planar surface in the 3D scene.
Based on this assumption, we associate a weight to the graph edge $(i, j)$, which quantifies our confidence about the two pixels $i$ and $j$ to belong to the same object.
Formally, first we define a ${B \times B}$ pixels search window centered at the pixel ${i}$. Then, for each pixel ${j}$ in the window we compute the following weight:
\begin{equation} \label{eq:graph_weights}
    w_{ij} =
    \exp \left( - \frac{ \Vert \bm{Q}_{i} - \bm{Q}_{j} \Vert_{F}^{2}}
    {2 \sigma_{int}^{2}} \right)
    \exp \left( - \frac{ \Vert i - j \Vert_{2}^{2}}
    {2 \sigma_{spa}^{2}} \right),
\end{equation}
where $\bm{Q}_{i} \in \mathbb{R}^{Q \times Q}$ is a patch centered at the pixel $i$ of the image $I$, $\| \cdot \|_{F}$ denotes the Frobenius norm and $\sigma_{int}$, $\sigma_{spa} \in \mathbb{R}_{> 0}$ are tunable parameters.
The first exponential in Eq.~\eqref{eq:graph_weights} has a high weight, hence high likelihood, when the values of the image pixels in two patches centered at $i$ and $j$ are similar; it is low otherwise  \cite{buades_a_review_2005, krahenb_efficient_2011, foi_foveated_2012, rossi_nonsmooth_2018}.
The second exponential then makes the weight decay as the Euclidean distance between $i$ and $j$ increases.

After the weights associated to all the pixels in the considered $B \times B$ search window have been computed, we design a \textit{$K$ Nearest Neighbours} graph by keeping only the $K \in \mathbb{N}$ edges with the largest weights.
Limiting the number of connections at each pixel to $K$ reduces the computation during the minimization of the problem in Eq.~\eqref{eq:problem_formulation}, on the one hand, and it avoids weak edges that may connect pixels belonging to different objects, on the other one.

%%%%%%%%%%%%%%%%%%%%%%%%%%%%%%%%%%%%%%%%%%%%%%%%%%%%%%%%%%%%%%%%%%%%%%%%%%%%%%%%%%%%%%%%%%%%%%%%%%%%
%%%%%%%%%%%%%%%%%%%%%%%%%%%%%%%%%%%%%%%%%%%%%%%%%%%%%%%%%%%%%%%%%%%%%%%%%%%%%%%%%%%%%%%%%%%%%%%%%%%%
\subsection{Solver} \label{subsec:solver}
The problem in Eq.~\eqref{eq:problem_formulation} is convex, but non smooth.
Multiple solvers specifically tailored for this class of problems exist, such as the \textit{Forward Backward Primal Dual (FBPD)} solver \cite{condat_primaldual_2013}.
However, the convergence of these methods calls for the estimation of multiple parameters before the actual minimization takes place, such as the operator norm of the implicit linear operator associated to the regularization term in Eqs.~\eqref{eq:prior_planes}--\eqref{eq:prior_normals}, which can be very time demanding.
Therefore, we decide to solve the problem in Eq.~\eqref{eq:problem_formulation} using \textit{Gradient Descent with momentum}, in particular \textit{ADAM} \cite{kingma_adam_2015}, as we empirically found it to be considerably faster (time-wise) than FBPD in our scenario.

Overall, our algorithm consists of two tasks: the graph construction and the solution of the problem in Eq.~\eqref{eq:problem_formulation} with ADAM.
Resorting to \cite{darbon_fast_2008}, the graph construction has a complexity $O(H W B \log_{2} B)$ regardless of the patch size $Q$.
The complexity of a single ADAM iteration is $O(H W K)$, with $K \ll B^{2}$, and it is due to the gradient computation.

Finally, to further speed up ADAM convergence, we adopt a multi-scale approach.
The noisy and possibly incomplete inverse depth map $\bar{d}$ is progressively down-sampled by a factor $r \in \mathbb{N}$ to get $\bar{d}^{\ell} \in \mathbb{R}^{\lfloor H/r^{\ell} \rfloor \times \lfloor W/r^{\ell} \rfloor}$ with $\ell = 0, \ldots, L-1$ and $L \in \mathbb{N}$ the number of scales.
An instance of the problem in Eq.~\eqref{eq:problem_formulation} is solved for each $\bar{d}^{\ell}$ and the solution at the scale ${\ell}$ is up-sampled and scaled by a factor $r$ to initialize the solver at the scale $\ell - 1$.\footnote{All up-sampling and down-sampling operations are performed using nearest neighbor interpolation.}
The scaling is a consequence of the relation $u^{\ell - 1} = r^{-1} u^{\ell}$.
In fact, the up/down-sampling operations emulate a change of the pixel area, while the camera sensor area remains constant.
We refer to the supplementary material for a formal proof.

%%%%%%%%%%%%%%%%%%%%%%%%%%%%%%%%%%%%%%%%%%%%%%%%%%%%%%%%%%%%%%%%%%
%%%%%%%%%%%%%%%%%%%%%%%%%%%%%%%%%%%%%%%%%%%%%%%%%%%%%%%%%%%%%%%%%%
%%%%%%%%%%%%%%%%%%%%%%%%%%%%%%%%%%%%%%%%%%%%%%%%%%%%%%%%%%%%%%%%%%
%%%%%%%%%%%%%%%%%%%%%%%%%%%%%%%%%%%%%%%%%%%%%%%%%%%%%%%%%%%%%%%%%%
% Experiments
%
\section{Experimental results} \label{sec:experiments}

We test the effectiveness of our joint depth refinement and normal estimation framework on the training splits of the \textit{Middlebury v3} stereo dataset \cite{scharstein_high_2014} at quarter resolution, of the \textit{KITTI 2015} stereo dataset \cite{menze_object_2015} and of the \textit{ETH3D} \textit{Multi-View Stereo (MVS)} dataset \cite{schops_multiview_2017} at half resolution.
Since these datasets come with ground truth depth maps but lack ground truth normals, we provide numerical results for the depth refinement part of the framework, while we provide only visual results for the normal estimation part.

Regarding the ground truth normal map $n_{gt}$, we approximate it by solving the system in Eq.~\eqref{eq:normal_sys} with $u = \nabla d_{gt}$, where the gradient is computed using a $5 \times 5$ pixels Gaussian derivative kernel with standard deviation $\sigma = 0.2$ pixels.
The small standard deviation permits to recover fine details, as the ground truth inverse depth map $d_{gt}$ is not affected by noise.
Although this does not permit a numerical evaluation, it permits to appreciate the normals estimated by our framework.

%%%%%%%%%%%%%%%%%%%%%%%%%%%%%%%%%%%%%%%%%%%%%%%%%%%%%%%%%%%%%%%%%%%%%%%%%%%%%%%%%%%%%%%%%%%%%%%%%%%%
%%%%%%%%%%%%%%%%%%%%%%%%%%%%%%%%%%%%%%%%%%%%%%%%%%%%%%%%%%%%%%%%%%%%%%%%%%%%%%%%%%%%%%%%%%%%%%%%%%%%
%%%%%%%%%%%%%%%%%%%%%%%%%%%%%%%%%%%%%%%%%%%%%%%%%%%%%%%%%%%%%%%%%%%%%%%%%%%%%%%%%%%%%%%%%%%%%%%%%%%%
\subsection{Middlebury and KITTI datasets}

Similarly to the recent disparity refinement method in \cite{tosi_leveraging_2019}, we refine the disparity maps computed via \textit{Semi-Global Matching (SGM)} \cite{hirschmuller_stereo_2008} and census-based \textit{Block Matching (BM)} \cite{zabih_non_param_1994}.
We compare our framework to the disparity refinement method recently proposed in \cite{tosi_leveraging_2019}, as it also relies on a confidence map and, most importantly, it showed to outperform many other widely used disparity refinement methods, e.g., \cite{perreault_median_2007, matoccia_locally_2009, ma_constant_2013, zhang_100_2014, barron_fast_2016}, on both the Middlebury and the KITTI datasets.
Moreover, since our new regularization in Eqs.~\eqref{eq:prior_planes}--\eqref{eq:prior_normals} resembles NLTGV \cite{ranftl_non_local_2014}, we compare to NLTGV as well.
In particular, we replace ${g(\cdot)}$ with NLTGV in our problem formulation in Eq.~\eqref{eq:problem_formulation}.

It is crucial to observe that, originally, NLTGV was introduced in the context of optical flow \cite{ranftl_non_local_2014} as a general purpose regularization, without any ambition to connect the geometry of the optical flow and the geometry of the underneath scene.
Here instead, we aim at modeling the joint piece-wise planarity of the inverse depth map and of the underneath scene explicitly.
In fact, the mixed $\ell_{1, 2}$--norms employed in both the terms of our regularization, as opposed to the simple $\ell_{1}$--norm of NLTGV, are chosen to make our regularization more robust in its global plane fitting.\footnote{A through analysis of the differences between the proposed regularization and NLTGV is provided in the supplementary material.}

The SGM and BM disparity maps to refine are provided by the authors in \cite{tosi_leveraging_2019}, who provided also their refined disparity maps and binary confidence maps.
In order to carry out a fair comparison, these confidence maps are used by all the methods considered in the experiments.
As described in \cite{tosi_leveraging_2019}, the considered binary confidence maps are the result of a learning-based framework trained on a split of the \textit{KITTI 2012} stereo dataset \cite{geiger_object_2012}, therefore there is no bias toward the Middlebury and KITTI 2015 datasets.

Since our framework assumes a depth map at its input, we convert the disparity map to be refined into a depth map and we then convert the refined depth map back to the disparity domain, in order to carry out the numerical evaluation.
The evaluation involves the \textit{bad pixel} metric, which is the percentage of pixels with an error larger than a predefined disparity threshold, together with the \textit{average absolute error (avgerr)} and the \textit{root mean square error (rms)}.
We carry out the evaluation on all the pixels with an available ground truth, regardless of the occlusions.

For a fair comparison, in the graph construction we adopt the same parameters for both NLTGV and our framework: weight parameters $\sigma_{int} = 0.07$ and $\sigma_{spa} = 3$ pixels, search window size ${B = 9}$, patch size ${P = 3}$ and maximum number of per pixel connections ${K = 20}$.
For both, we also set the number of scales $L = 2$ and $r = 2$.
Instead, the multipliers $\lambda$ and $\alpha$ in front and inside the regularization $g(\cdot)$, respectively, are the result of a grid search and listed below.

%%%%%%%%%%%%%%%%%%%%%%%%%%%%%%%
\paragraph{Middlebury dataset}
%%%%%%%%%%%%%%%%%%%%%%%%%%%%%%%

The Middlebury training dataset \cite{scharstein_high_2014} consists of $15$ indoor scenes carefully crafted to challenge modern stereo algorithms.
Some scenes contain multiple untextured planar surfaces, which represent a hard challenge for stereo methods but are compliant with the model underneath our framework; other scenes are inherently non piece-wise planar instead.
Due to its variety, the Middlebury dataset permits to evaluate the flexibility of our framework to different settings.

For NLTGV we set $\lambda = 7.5$ and $\alpha = 50$, regardless of the scale.
For our framework and SGM disparity maps at the input, we set $\lambda = 15$ and $25$ at the low and high scales, respectively; for BM disparity maps at the input instead, we set $\lambda = 10$ and $20$ at the low and high scales, respectively; we set $\alpha = 3.5$ regardless of the input disparity map.

The results of our experiments on the Middlebury dataset are presented in Table~\ref{tab:middlebury_all}.
When BM is considered, our framework outperforms the method in \cite{tosi_leveraging_2019} and NLTGV in all the considered metrics.
Similarly, when SGM is considered, our framework outperforms the method in \cite{tosi_leveraging_2019} and NLTGV in four of the five metrics; in the \textit{bad 1px} metric, where the best error is achieved by the method in \cite{tosi_leveraging_2019}, our result is comparable.
Moreover, in the most common \textit{bad 2px} metric, our framework always provides the best error regardless of the input disparity map.
Clearly, some scenes in the dataset are far from fulfilling our piece-wise planar assumption, e.g., \texttt{Jadeplant} and \texttt{Pipes}: these affect the average results in Table~\ref{tab:middlebury_all} and mask the large improvement exhibited by our framework in those scenes which fulfill the assumption even partially.

\begin{table}
    \caption{Disparity refinement on the Middlebury dataset \cite{scharstein_high_2014}.
    The first column specifies the stereo method whose disparity map is refined.
    The second column provides the error metric used in the evaluation: \textit{bad px} thresholds, the \textit{average absolute error (avgerr)} and the \textit{root mean square error (rms)}.
    All the pixels with a ground truth disparity are considered.
    The columns from four to six report the error of the disparity maps refined by the method in \cite{tosi_leveraging_2019}, NLTGV \cite{ranftl_non_local_2014}, our method.
    The best result for each row is in bold.
    \vspace{5pt}
    }
    \centering
    \fontsize{8}{9}
    \selectfont
    \begin{tabular}{|l|c|cccg|}
        \hline
        & Err. metric & Input & \cite{tosi_leveraging_2019} & \cite{ranftl_non_local_2014} & $\text{Ours}$ \\
        \hline
        \multirow{5}{*}{SGM \cite{hirschmuller_stereo_2008}}
            & bad 0.5px & 41.33 & 39.14 & 36.57 & \textbf{35.70} \\
            & bad 1px & 28.90 & \textbf{25.58} & 26.02 & {25.71} \\
            & bad 2px & 23.48 & 19.55 & 19.88 & \textbf{19.25} \\
            & avgerr & 4.06 & 3.32 & 3.31 & \textbf{2.87} \\
            &  rms & 9.75 & 8.27 & 7.99 & \textbf{6.86} \\
        \hline
        \multirow{5}{*}{BM \cite{zabih_non_param_1994}}
            & bad 0.5px & 47.48 & 39.01 & 38.49 & \textbf{35.01} \\
            & bad 1px & 37.56 & 25.83 & 28.28 & \textbf{25.40} \\
            & bad 2px & 33.98 & 20.61 & 22.03 & \textbf{19.41} \\
            & avgerr & 8.41 & 3.48 & 3.35 & \textbf{2.79} \\
            &  rms & 17.32 & 8.58 & 7.91 & \textbf{6.97} \\
        \hline
    \end{tabular}
    \label{tab:middlebury_all}
\end{table}

\begin{figure}
    \centering
    \setlength{\tabcolsep}{1pt}
    \renewcommand{\arraystretch}{0.1}
    \begin{tabular}{cccc}
    	\rotatebox{90}{\makebox[1.8cm] {\scriptsize Reference}} &
    	\includegraphics[width=0.15\textwidth]{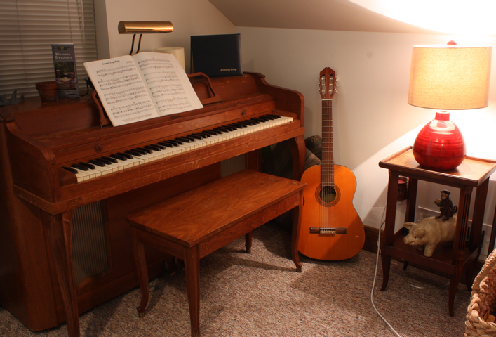} &
    	\includegraphics[width=0.15\textwidth]{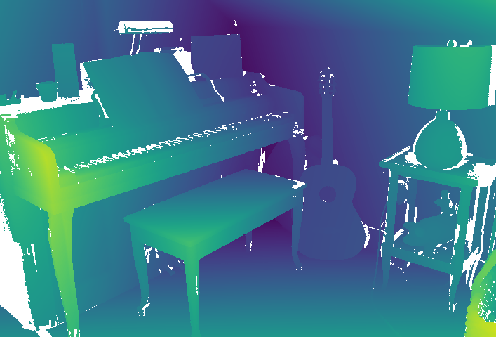} &
	\includegraphics[width=0.15\textwidth]{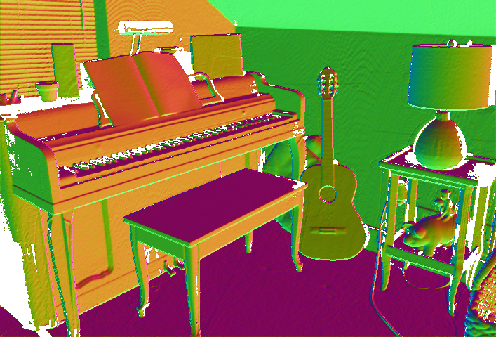} \\
	\\
    	\rotatebox{90}{\makebox[1.8cm] {\scriptsize BM \cite{zabih_non_param_1994}}} &
    	\includegraphics[width=0.15\textwidth]{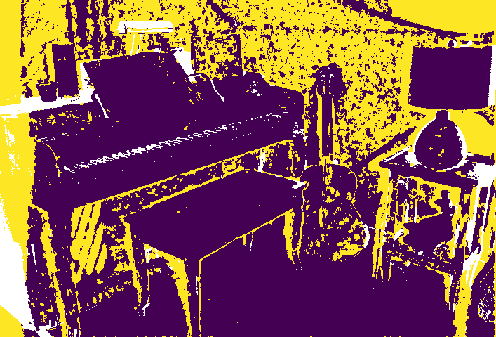} &
    	\begin{overpic}[width=0.15\textwidth]{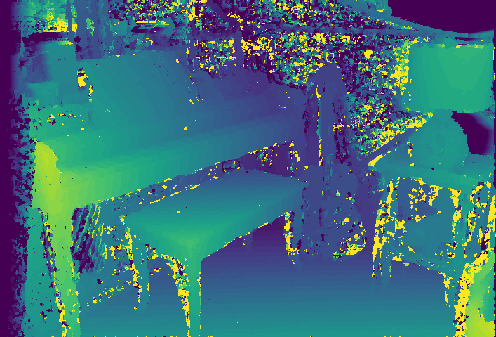}
            \put(68, 4){\color{white} \scriptsize \textbf{31.88\%}}
    	\end{overpic} &
    	\includegraphics[width=0.15\textwidth]{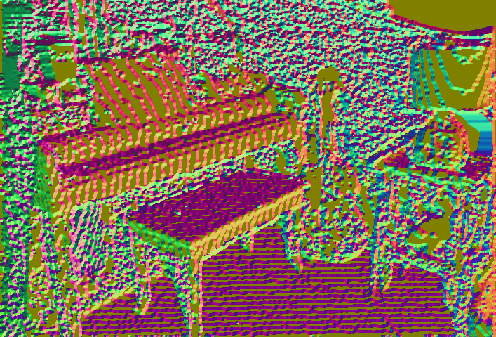} \\
    	\\
    	\rotatebox{90}{\makebox[1.8cm] {\scriptsize \cite{tosi_leveraging_2019}}} &
    	\includegraphics[width=0.15\textwidth]{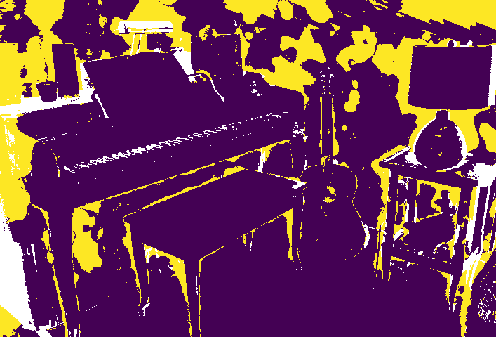} &	
    	\begin{overpic}[width=0.15\textwidth]{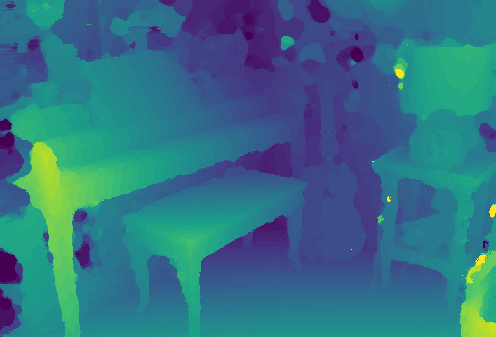}
    	    \put(68, 4){\color{white} \scriptsize \textbf{18.29\%}}
    	\end{overpic} &
    	\includegraphics[width=0.15\textwidth]{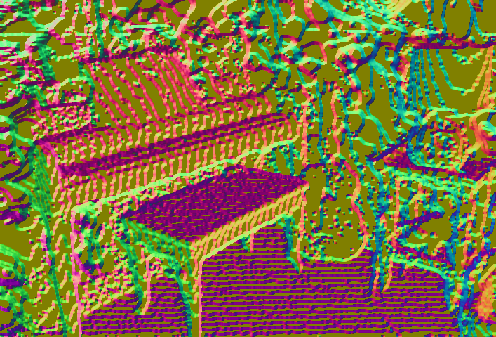} \\
    	\\
    	\rotatebox{90}{\makebox[1.8cm] {\scriptsize NLTGV \cite{ranftl_non_local_2014}}} &
    	\includegraphics[width=0.15\textwidth]{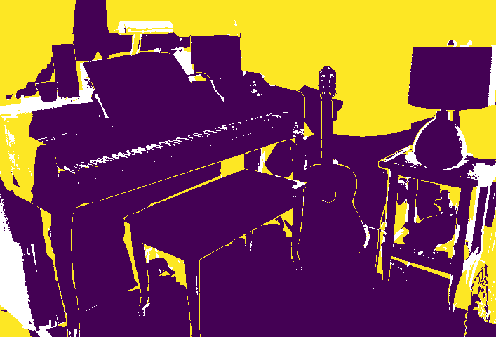} &
    	\begin{overpic}[width=0.15\textwidth]{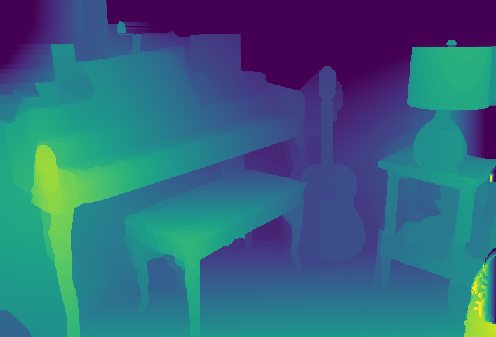}
    	    \put(68, 4){\color{white} \scriptsize \textbf{27.86\%}}
    	\end{overpic} &
    	\includegraphics[width=0.15\textwidth]{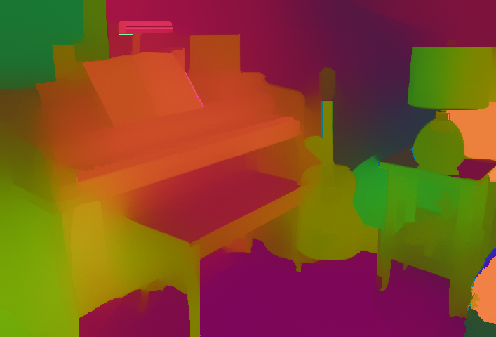} \\
    	\\
    	\rotatebox{90}{\makebox[1.8cm] {\scriptsize $\text{Ours}$}} &
        	\includegraphics[width=0.15\textwidth]{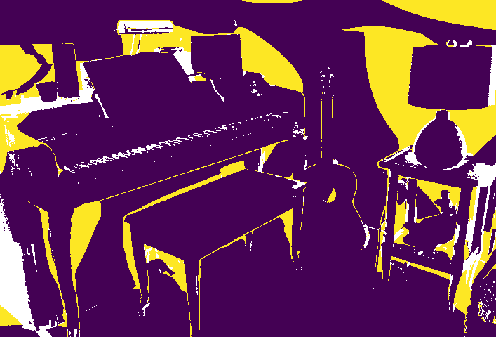} &
    	\begin{overpic}[width=0.15\textwidth]{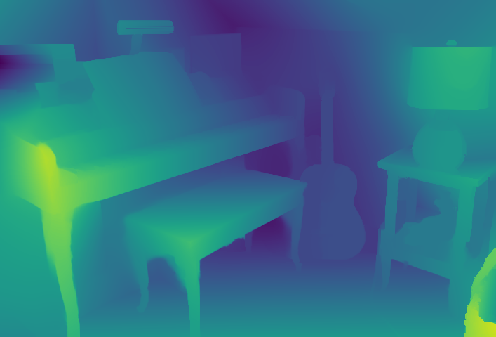}
    	    \put(68, 4){\color{white} \scriptsize \textbf{15.25\%}}
    	\end{overpic} &
    	\includegraphics[width=0.15\textwidth]{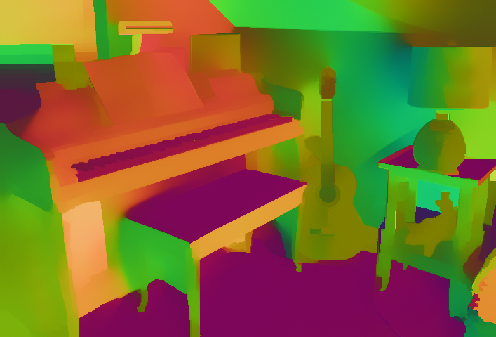} \\[5pt]
    \end{tabular}
    \caption{Middlebury \cite{scharstein_high_2014} scene \texttt{Piano}.
    	The first row hosts, from left to right, the reference image and the ground truth disparity and normal maps.
    	Each other row hosts, from left to right, the \textit{bad 2px} disparity error mask and the disparity and normal map.
    	The second row refers to the stereo method BM \cite{zabih_non_param_1994}, whose disparity is refined by the method in \cite{tosi_leveraging_2019}, NLTGV \cite{ranftl_non_local_2014} and ours, in the rows three to five, respectively.
    	The pixels in the error maps are color coded: error within $2$px in dark blue, error larger than $2$px in yellow, missing ground truth in white.
    	The \textit{bad 2px} error percentage is reported on the bottom right corner of each disparity map.}
    \label{fig:middlebury_piano_bm}
\end{figure}

\begin{figure*}[t!]
    \centering
    \setlength{\tabcolsep}{1pt}
    \renewcommand{\arraystretch}{0.1}
    \begin{tabular}{ccccc}
    \scriptsize{Reference} &
    \scriptsize{BM \cite{zabih_non_param_1994}} &
    \scriptsize{\cite{tosi_leveraging_2019}} &
    \scriptsize{NLTGV \cite{ranftl_non_local_2014}} &
    \scriptsize{Ours} \\
    \\
    \includegraphics[width=0.195\textwidth]{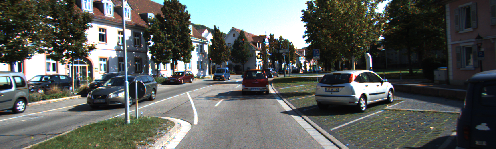} &
    \includegraphics[width=0.195\textwidth]{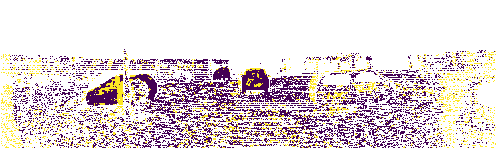} &
    \includegraphics[width=0.195\textwidth]{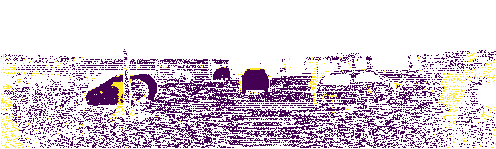} &
    \includegraphics[width=0.195\textwidth]{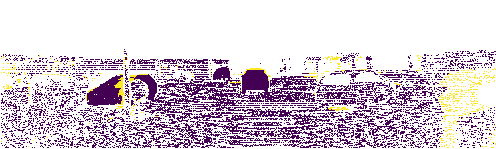} &
    \includegraphics[width=0.195\textwidth]{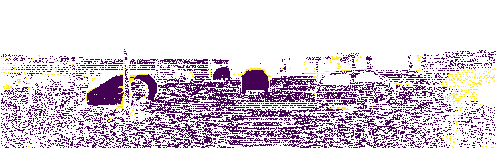} \\
    \\
    \includegraphics[width=0.195\textwidth]{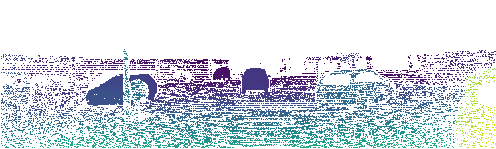} &
    \begin{overpic}[width=0.195\textwidth]{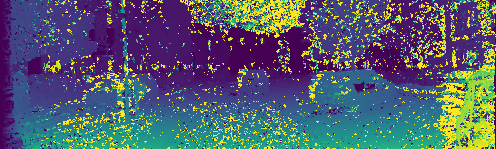}
       \put(74, 2){\color{white} \scriptsize \textbf{26.97\%}}
    \end{overpic} &
    \begin{overpic}[width=0.195\textwidth]{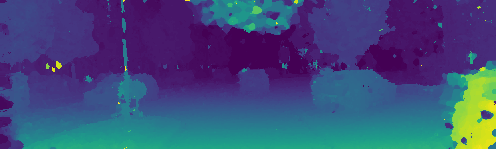}
        \put(74, 2){\color{white} \scriptsize \textbf{\phantom{0}7.99\%}}
    \end{overpic} &
    \begin{overpic}[width=0.195\textwidth]{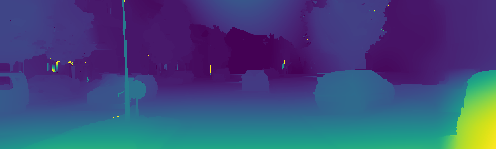}
        \put(74, 2){\color{white} \scriptsize \textbf{\phantom{0}7.39\%}}
    \end{overpic} &
    \begin{overpic}[width=0.195\textwidth]{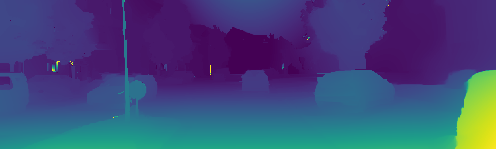}
        \put(74, 2){\color{white} \scriptsize \textbf{\phantom{0}5.46\%}}
    \end{overpic} \\
    \\
    \includegraphics[width=0.195\textwidth]{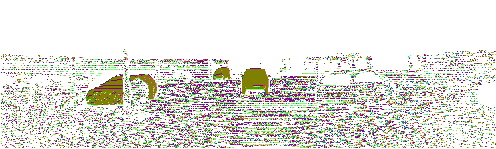} &
    \includegraphics[width=0.195\textwidth]{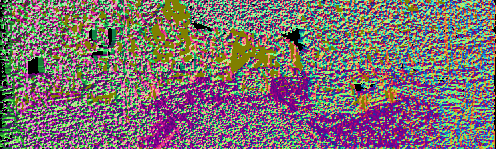} &
    \includegraphics[width=0.195\textwidth]{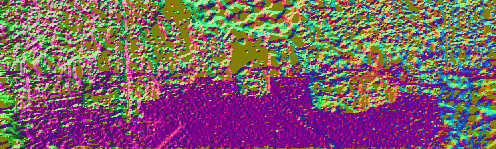} &
    \includegraphics[width=0.195\textwidth]{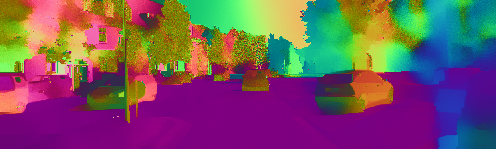} &
    \includegraphics[width=0.195\textwidth]{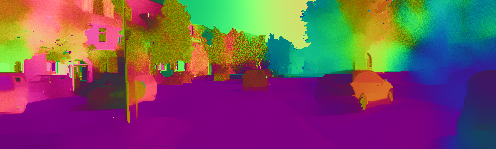} \\[5pt]
    \end{tabular}
    \caption{KITTI \cite{menze_object_2015} scene \texttt{126}.
    The first column hosts, from top to bottom, the reference image and the ground truth disparity and normal maps.
    Each other column hosts, from top to bottom, the \textit{bad 3px} disparity error mask and the disparity and normal map.
    The second column refers to the stereo method BM \cite{zabih_non_param_1994}, whose disparity is refined by the method in \cite{tosi_leveraging_2019}, NLTGV \cite{ranftl_non_local_2014} and ours, in the columns three, four and five, respectively.
    The pixels in the error maps are color coded: error within $3$px in blue, error larger than $3$px in yellow, missing ground truth in white.
    The \textit{bad 3px} error percentage is reported on the bottom right corner of each disparity map.}
    \label{fig:kitti_126_bm}
\end{figure*}

In Figure~\ref{fig:middlebury_piano_bm} we provide the results of our experiments on the scene \texttt{Piano}, when the stereo method BM is considered.
The normal map associated to the input BM disparity map and to the one refined by the method in \cite{tosi_leveraging_2019} are computed with the same approach adopted for the ground truth normal map, while employing $\sigma = 5$ pixels in order to handle the noise.
In fact, the input BM disparity is significantly noisy, especially in the walls surrounding the piano.
The method in \cite{tosi_leveraging_2019} manages to decrease the error in some areas of the surrounding walls: however, since no global consistency is considered, the result is a speckled error.
Instead, our method manages to approximate the surrounding walls better, using multiple planes.
Finally, NLTGV fails to capture the geometry of the surrounding wall, as its relying on a simple $\ell_{1}$--norm makes it more sensible to outliers than our mixed $\ell_{1, 2}$--norm when trying to fit planes.

%%%%%%%%%%%%%%%%%%%%%%%%%%%%%%%
\paragraph{KITTI dataset}
%%%%%%%%%%%%%%%%%%%%%%%%%%%%%%%

The KITTI 2015 training dataset \cite{menze_object_2015} consists of $200$ scenes captured from the top of a moving car.
As a consequence, the prevalent content of each scene are the road, possible vehicles and possible buildings at the two sides of the road.
At a first glance, this content may seem to match our piece-wise planar assumption.
In practice, however, the buildings at the sides of the road are mostly occluded by the vegetation, which is far from piece-wise planar.
We select $20$ scenes randomly and test our framework on them, in order to analyze its flexibility.

For NLTGV we set $\lambda = 7.5$ and $\alpha = 15$ regardless of the scale.
For our framework we set $\lambda = 10$ and $20$ at the lowest and highest scales, respectively, while we set $\alpha = 15$ regardless of the scale.

The results of our experiments on the KITTI dataset are presented in Table~\ref{tab:kitti_all}.
Regardless of the considered metric and input stereo method, NLTGV outperforms the method in \cite{tosi_leveraging_2019}, while our framework
outperforms all the others.
Moreover, when the most common \textit{bad 3px} error is considered, our framework improves the SGM and BM disparity maps by more than $4.57\%$ and $31.75\%$, respectively.

In Figure~\ref{fig:kitti_126_bm} we provide the results of our experiments on the scene \texttt{126}, when the stereo method BM is considered.
The method in \cite{tosi_leveraging_2019}, NLTGV and our framework manage all to reduce sensibly the high amount of noise that affects the input disparity map, represented by the yellow speckles.
However, only NLTGV and our framework manage to preserve fine details like the pole on the left side of the image, which appears broken in the disparity map associated to \cite{tosi_leveraging_2019}.
Finally, our framework provides the sharpest disparity map, as NLTGV exhibits some disparity bleeding at object boundaries.
This is visible on the car at the bottom right corner of the image, both by observing the disparity maps and the error masks.
This is also confirmed by the numerical results, as our \textit{bad 3px} error is significantly lower.

\begin{table}[t!]
    \caption{Disparity refinement on the KITTI dataset \cite{menze_object_2015}.
    The first column specifies the stereo method whose disparity map is refined.
    The second column specifies the considered error metric: \textit{bad px} thresholds, the \textit{average absolute error (avgerr)} and the \textit{root mean square error (rms)}.
    All the pixels with a ground truth disparity are considered.
    The columns from four to six report the error of the disparity maps refined by the method in \cite{tosi_leveraging_2019}, NLTGV \cite{ranftl_non_local_2014} and our method, respectively.
    The best result for each row is in bold.
    \vspace{5pt}
    }
    \centering
    \fontsize{8}{9}
    \selectfont
    \begin{tabular}{|l|c|cccg|}
        \hline
        & Err. metric & Input & \cite{tosi_leveraging_2019} & \cite{ranftl_non_local_2014} & Ours \\
        \hline
        \multirow{4}{*}{SGM \cite{hirschmuller_stereo_2008}}
            & bad 2px & 14.25 & 11.58 & 10.49 & \textbf{9.82} \\
            & bad 3px & 10.11 & 7.65 & 6.07 & \textbf{5.54} \\
            & avgerr & 21.12 & 2.97 & 1.62 & \textbf{1.51} \\
            & rms & 46.50 & 8.49 & 7.91 & \textbf{7.88} \\
        \hline
        \multirow{4}{*}{BM \cite{zabih_non_param_1994}}
            & bad 2px & 40.96 & 16.75 & 11.09 & \textbf{10.54} \\
            & bad 3px & 38.15 & 12.80 & 6.76 & \textbf{6.40} \\
            & avgerr & 1.94 & 1.63 & 1.30 & \textbf{1.22} \\
            & rms & 5.46 & 4.52 & 3.43 & \textbf{3.15} \\
        \hline
    \end{tabular}
    \label{tab:kitti_all}
\end{table}

%%%%%%%%%%%%%%%%%%%%%%%%%%%%%%%%%%%%%%%%%%%%%%%%%%%%%%%%%%%%%%%%%%%%%%%%%%%%%%%%%%%%%%%%%%%%%%%%%%%%
%%%%%%%%%%%%%%%%%%%%%%%%%%%%%%%%%%%%%%%%%%%%%%%%%%%%%%%%%%%%%%%%%%%%%%%%%%%%%%%%%%%%%%%%%%%%%%%%%%%%
%%%%%%%%%%%%%%%%%%%%%%%%%%%%%%%%%%%%%%%%%%%%%%%%%%%%%%%%%%%%%%%%%%%%%%%%%%%%%%%%%%%%%%%%%%%%%%%%%%%%
\subsection{ETH3D dataset}

Large scale 3D reconstruction methods \cite{galliani_massively_2015, schoenberger_pixelwise_2016, kuhn_tv_2017, huang_deepmvs_2018, yao_mvsnet_2018, xu_multi_2019, kuhn_deep_2020} estimate the depth map of a reference image from a large number of input images of the same scene, by leveraging geometric and photometric constraints, and subsequently fuse them to produce a model of the scene itself.
Large scale 3D reconstruction methods can largely benefit from a refinement of the estimated depth maps and can exploit the corresponding normal maps during the fusion step.
In order to demonstrate the suitability of our joint depth refinement and normal estimation framework on high resolution images from challenging MVS configurations, we test it on the training split of the ETH3D dataset \cite{schops_multiview_2017}, a popular benchmark for large scale 3D reconstruction algorithms, involving both indoor and outdoor sequences.

\begin{table}[t!]
    \caption{Refinement of MVS-derived \cite{kuhn_deep_2020} depth maps from the ETH3D training dataset \cite{schops_multiview_2017}.
    The table is divided into a top and a bottom sub-table, with their first columns specifying the test scenes, whose number of images is specified in brackets.
    The top sub-table reports the percentage of pixels with an error exceeding a predefined threshold: $2$cm and $5$cm.
    The bottom sub-table reports the \textit{average absolute error (avgerr)} and the \textit{root mean square error (rms)} in the second and third columns, respectively.
    For each scene and error metric, the best result is in bold.
    \vspace{5pt}
    }
    \centering
    \fontsize{8}{9}
    \selectfont
    \begin{tabular}{|c|ccg|ccg|}
        \hline
        &
        \multicolumn{3}{c|}{2cm} &
        \multicolumn{3}{c|}{5cm} \\
        &
        Input & \cite{ranftl_non_local_2014} & Ours &
        Input & \cite{ranftl_non_local_2014} & Ours \\
        \hline
        Pipes (14)  &
        18.16 & {11.17} & \textbf{10.71} &
        14.18 & 7.64 &  \textbf{7.10} \\
        {Delivery} (44) &
        24.15 & {19.20} & \textbf{18.33} &
        12.05 & {6.41}  & \textbf{5.80} \\
        Office (26) &
        47.54 & {39.34} & \textbf{38.59} &
        39.32 & 30.23 & \textbf{29.13} \\
        \hline
        avg. &
        29.95 & {23.24} & \textbf{22.54} &
        21.85 & 14.76 & \textbf{14.08} \\
        \hline
    \end{tabular}\\[1.5pt]
    \begin{tabular}{|c|ccg|ccg|}
        \hline
        &
        \multicolumn{3}{c|}{avgerr} &
        \multicolumn{3}{c|}{rms} \\
        &
        Input & \cite{ranftl_non_local_2014} & Ours &
        Input & \cite{ranftl_non_local_2014} & Ours \\
        \hline
        Pipes (14)  &
        0.347 & {0.119} & \textbf{0.082} &
        2.090 & 0.685 &  \textbf{0.460} \\
        {Delivery} (44) &
        0.233 & {0.025} & \textbf{0.023} &
        26.06 & {0.227} & \textbf{0.211} \\
        Office (26) &
        0.330 & {0.183} & \textbf{0.167} &
        1.218 & 0.409 & \textbf{0.381} \\
        \hline
        avg. &
        0.303 & {0.109} & \textbf{0.091} &
        9.303 & 0.440 & \textbf{0.351} \\
        \hline
    \end{tabular}
    \label{tab:eth3d}
\end{table}

The ground truth ETH3D depth maps are very sparse, but characterized by twice the resolution adopted in our tests.
Therefore, similarly to \cite{huang_deepmvs_2018}, we back project the sparse ground truth depth maps to half resolution in order to get denser ones, to be used in our evaluation.
In \cite{kuhn_deep_2020}, the authors propose a novel deep-network-based confidence prediction framework for depth maps computed by MVS algorithms, hence in the context of large baselines and severe occlusions.
For our experiments, in order to estimate the confidence map $m$, we re-train the network proposed in \cite{kuhn_deep_2020} jointly on the synthetic dataset proposed in the same work and on the dense ground truth depth maps of the ETH3D training split.
For an unbiased evaluation, we extract three sequences of the ETH3D training split (\texttt{Pipes}, \texttt{Office}, \texttt{Delivery Area}) from the training procedure and use them exclusively for our evaluation.
We compare our refined depth and normal maps against the depth and normal maps derived by the \textit{PatchMatch}-based \cite{bleyer_patchmatch_2011} MVS method presented in \cite{kuhn_deep_2020}.

For both NLTGV and our framework we set $\lambda = 7.5$ and $\alpha = 7.5$ regardless of the scale, adopt the graph parameters selected for Middlebury and KITTI, set the number of scales $L = 4$ and $r = 2$.
The continuous confidence map provided by the trained network is binarized with a $0.5$ threshold.

\begin{figure}
    \centering
    \setlength{\tabcolsep}{1pt}
    \renewcommand{\arraystretch}{0.1}
    \begin{tabular}{cccc}
    \rotatebox{90}{\makebox[1.8cm] {\scriptsize Reference}} &
	\includegraphics[width=0.31\columnwidth]{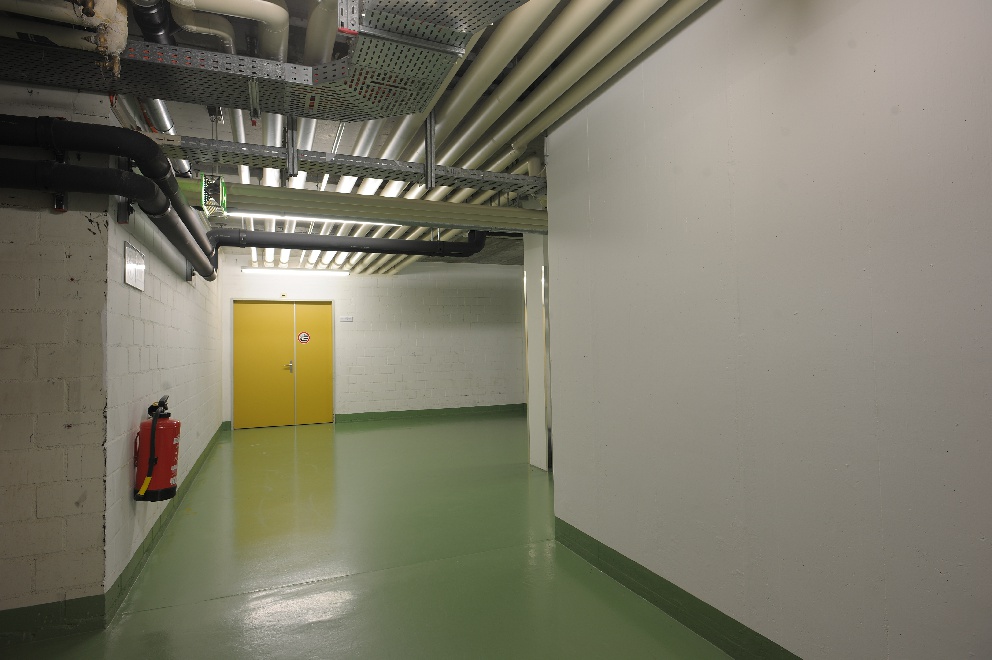} &
	\includegraphics[width=0.31\columnwidth]{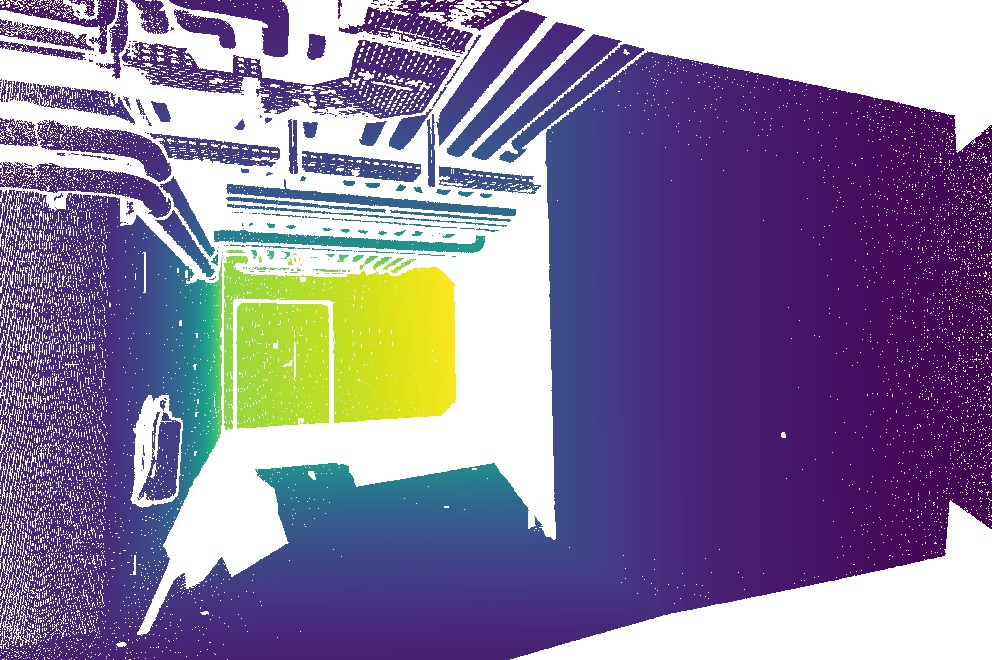} &
	\includegraphics[width=0.31\columnwidth]{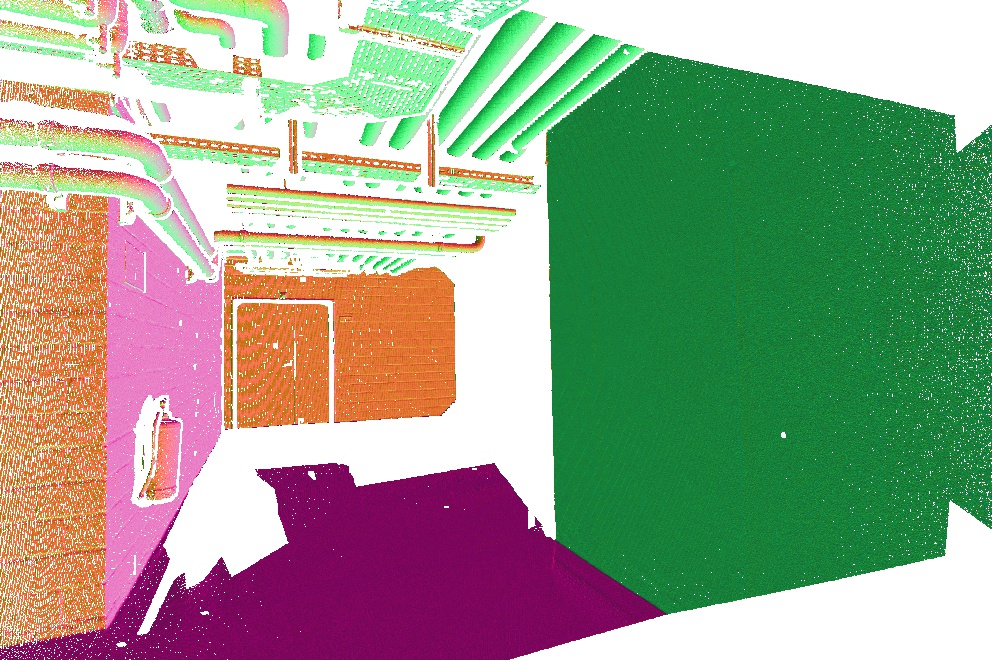} \\
	\rotatebox{90}{\makebox[1.8cm]{\scriptsize Input}} &
	\includegraphics[width=0.31\columnwidth]{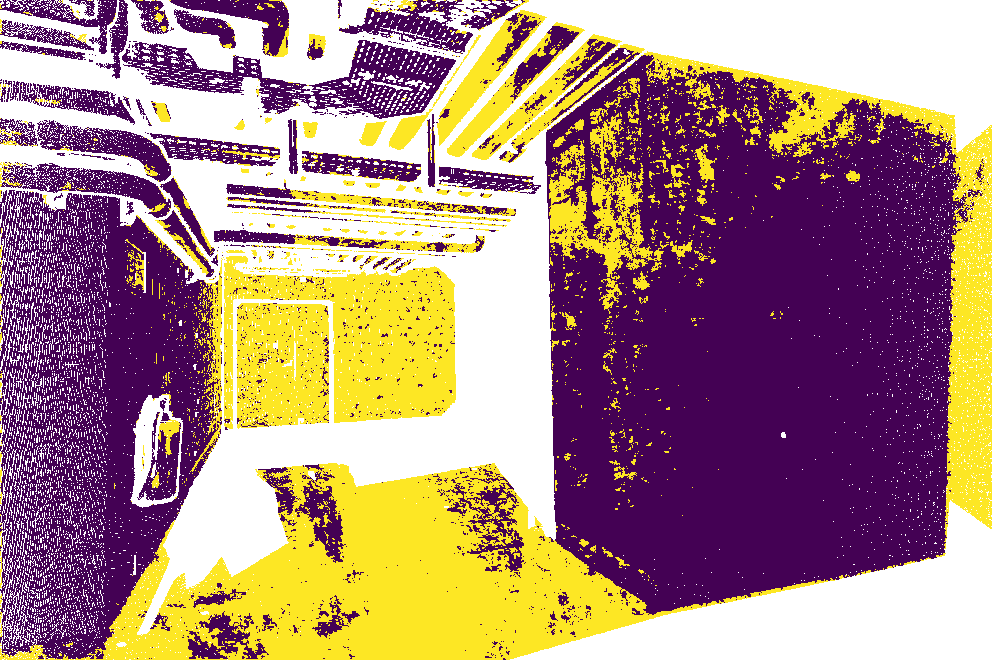} &
	\begin{overpic}[width=0.31\columnwidth]{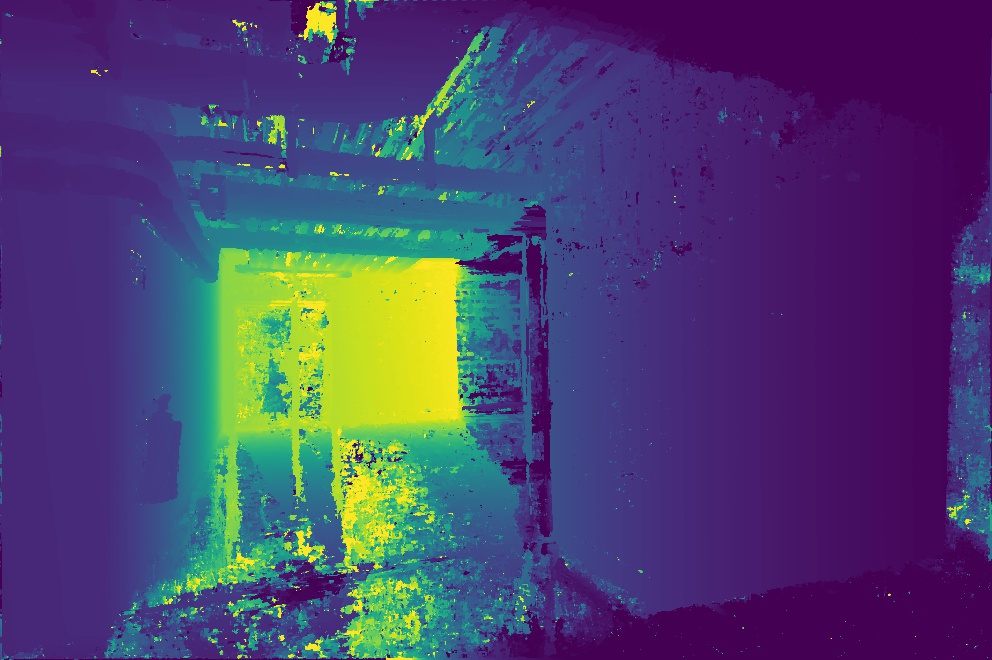}
	    \put(66, 4){\color{white} \scriptsize \textbf{33.22\%}}
	\end{overpic} &
	\includegraphics[width=0.31\columnwidth]{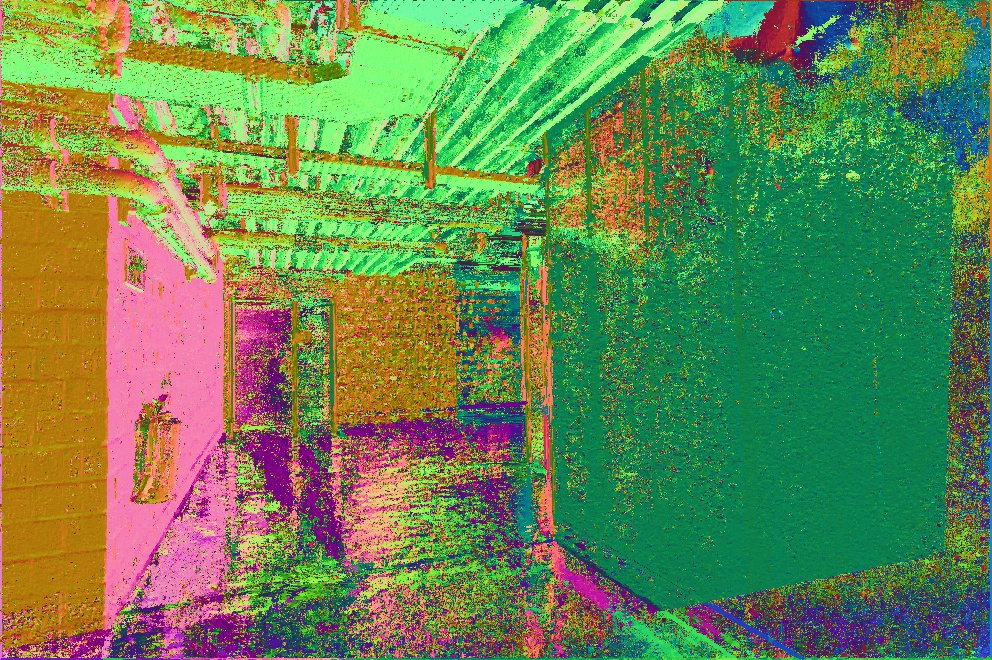} \\
	\rotatebox{90}{\makebox[1.8cm]{\scriptsize NLTGV \cite{ranftl_non_local_2014}}} &
	\includegraphics[width=0.31\columnwidth]{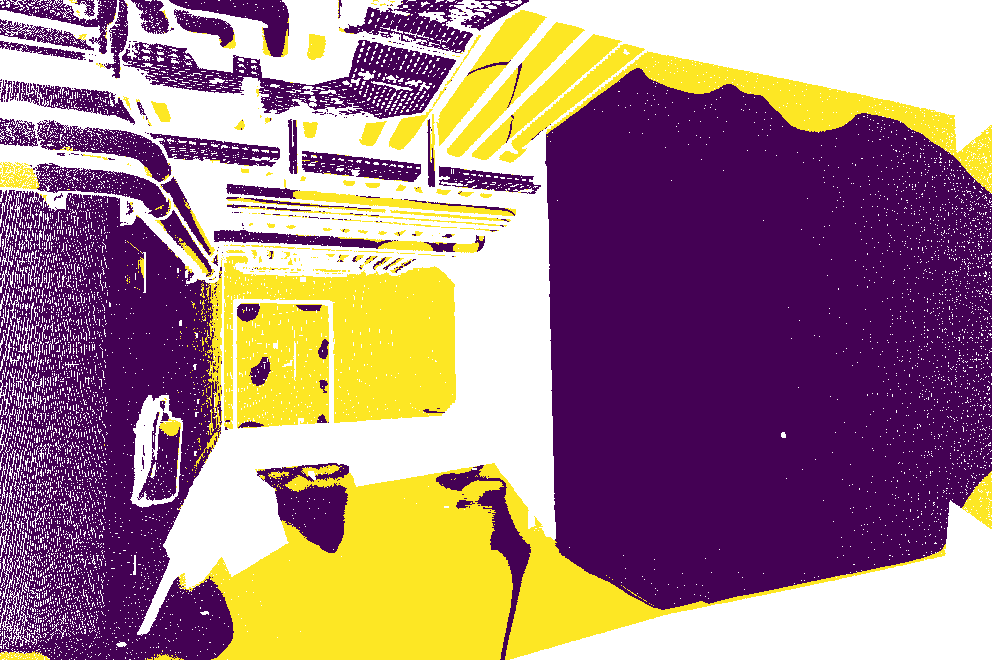} &
	\begin{overpic}[width=0.31\columnwidth]{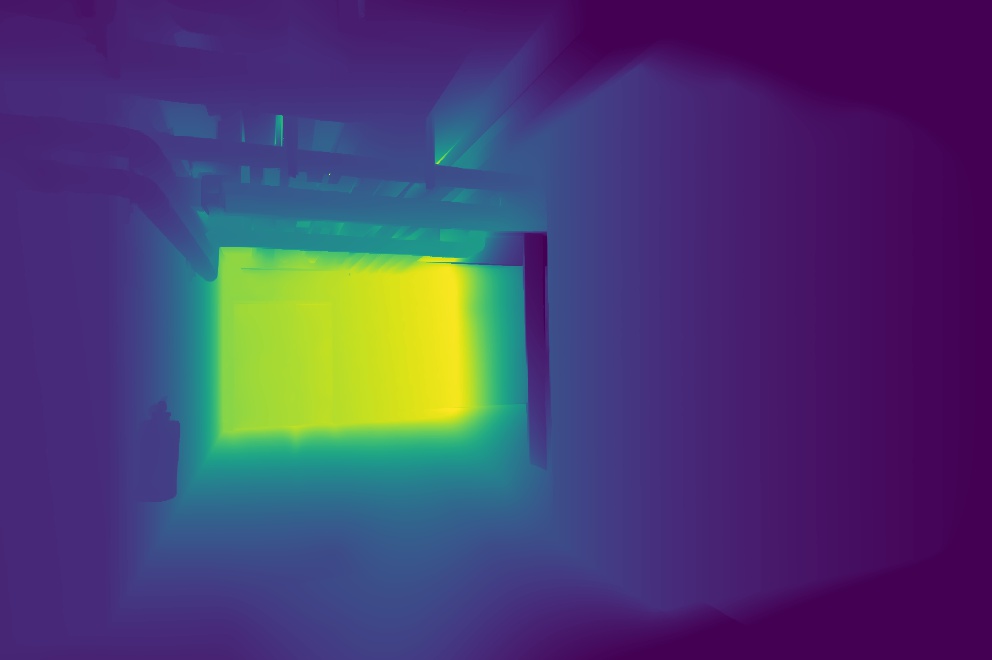}
	    \put(66, 4){\color{white} \scriptsize \textbf{24.34\%}}
	\end{overpic} &
	\includegraphics[width=0.31\columnwidth]{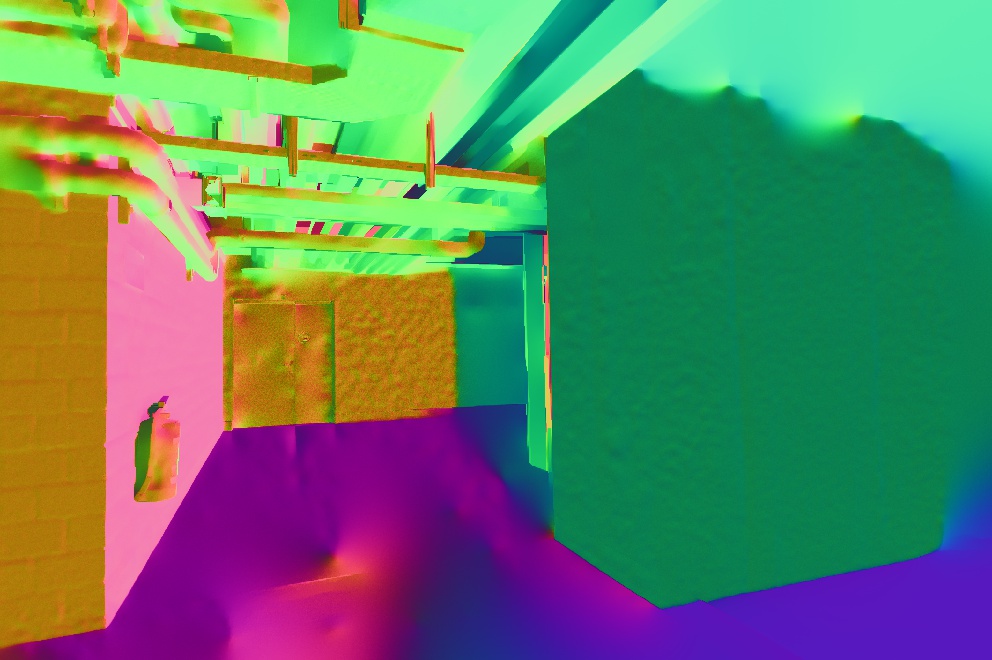} \\
	\rotatebox{90}{\makebox[1.8cm]{\scriptsize Ours}} &
	\includegraphics[width=0.31\columnwidth]{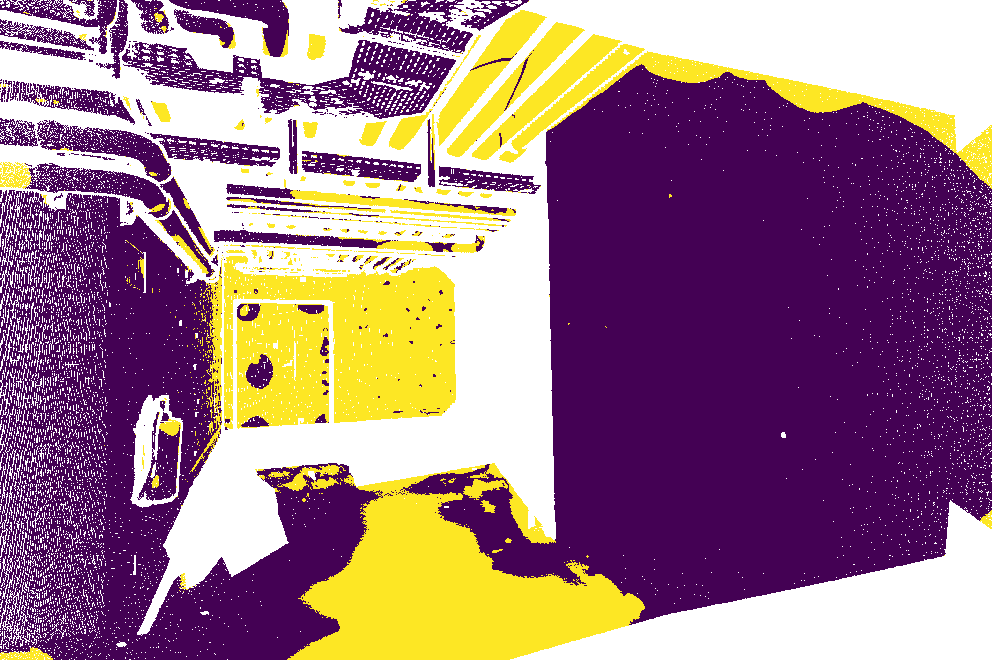} &
	\begin{overpic}[width=0.31\columnwidth]{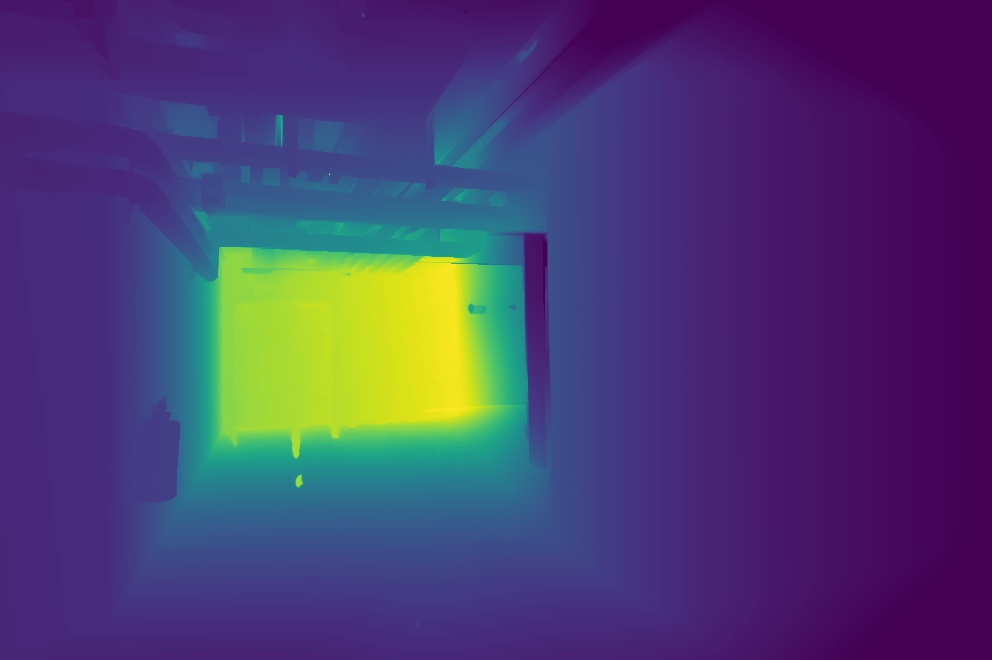}
	    \put(66, 4){\color{white} \scriptsize \textbf{22.20\%}}
	\end{overpic} &
	\includegraphics[width=0.31\columnwidth]{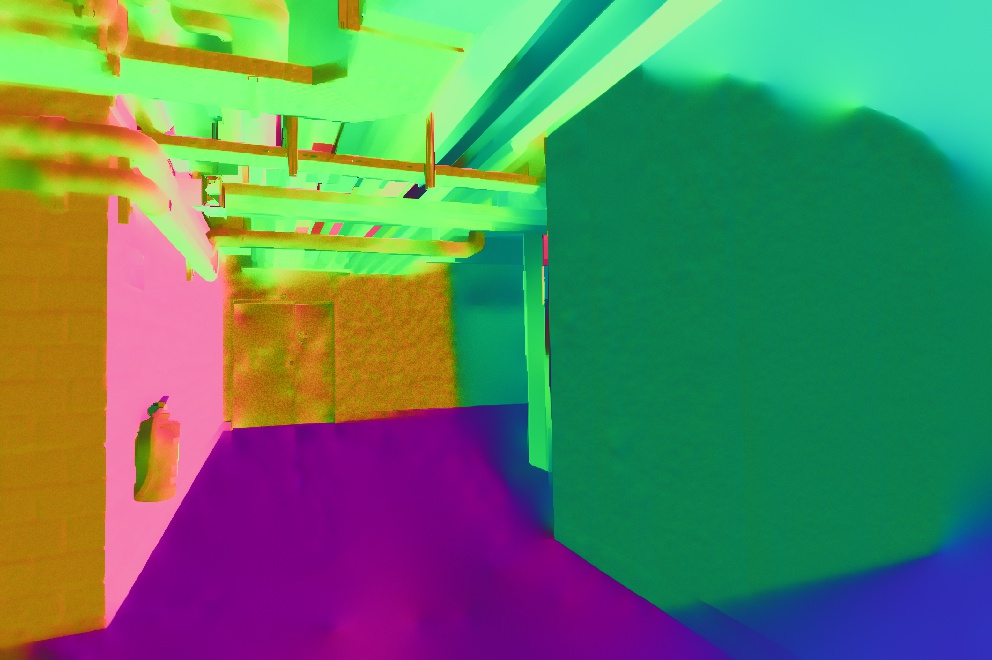} \\[5pt]
    \end{tabular}
    \caption{ETH3D \cite{schops_multiview_2017} \texttt{Pipes}.
    The first row hosts, from left to right, the reference image and the ground truth depth and normal maps.
    Each other row hosts, from left to right, the $2$cm error map, the depth and normal maps.
    The second row refers to the MVS method \cite{kuhn_deep_2020}, whose depth is refined by NLTGV \cite{ranftl_non_local_2014} and our method in the rows three and four, respectively.
    The pixels in the error maps are color coded: error within $2$cm in blue, error larger than $2$cm in yellow, missing ground truth in white.
    The error percentage associated to the $2$cm threshold is reported on the bottom right corner of each depth map.}
    \label{fig:eth3d_pipes}
\end{figure}

Table~\ref{tab:eth3d} compares the input MVS depth map with those refined by NLTGV and our method.
The top part of the table reports the percentage of pixel, computed over all the pixels of all the images in the sequence, with an error within a given threshold: $2$cm and $5$cm.
On average, our method outperforms NLTGV and manages to improve the input depth maps by more than $7\%$ when the $2$cm threshold, the most common in the ETH3D benchmark, is considered.
In the bottom part of the same table, we provide also the \textit{average absolute error (avgerr)} and the \textit{root mean square error (rms)}.
The \textit{rms} metric is very sensitive to outliers and, especially in the \texttt{Delivery Area} sequence, it highlights our improvement over the input depth map.
Finally, a visual example is provided in Figure~\ref{fig:eth3d_pipes} for the sequence \texttt{Pipes}, which is characterized by multiple untextured planar surfaces, representing a hard challenge for MVS methods.
Our method targets exactly these scenarios instead: in fact, it manages to refine the input depth map by capturing the main planes, as exemplified by the estimated normal map.
Moreover, it manages to fit better planes than NLTGV, which fails to capture the correct floor orientation.

%%%%%%%%%%%%%%%%%%%%%%%%%%%%%%%%%%%%%%%%%%%%%%%%%%%%%%%%%%%%%%%%%%
%%%%%%%%%%%%%%%%%%%%%%%%%%%%%%%%%%%%%%%%%%%%%%%%%%%%%%%%%%%%%%%%%%
%%%%%%%%%%%%%%%%%%%%%%%%%%%%%%%%%%%%%%%%%%%%%%%%%%%%%%%%%%%%%%%%%%
%%%%%%%%%%%%%%%%%%%%%%%%%%%%%%%%%%%%%%%%%%%%%%%%%%%%%%%%%%%%%%%%%%
% Conclusions
%
\section{Conclusions} \label{sec:conclusions}

In this article, we presented a variational framework to address the problem of depth map refinement.
In particular, we cast the problem into the minimization of a cost function involving a data fidelity and a graph-based regularization term.
The latter enforces piece-wise planar solutions explicitly,
by estimating the depth map and the corresponding normal map jointly,
as most human made environments exhibit a planar bias.
Moreover, the graph-based nature of the regularization makes our framework flexible enough to handle non fully piece-wise planar scenes as well.
We showed that the proposed framework outperforms state-of-the-art depth refinement methods when the considered scene meets our piece-wise planar assumption, and it leads to competitive results otherwise.
Interesting perspectives include the a priori segmentation of the reference image into \textit{planar} and \textit{non planar} areas, so that the strength of the regularization could be adapted accordingly.

%%%%%%%%%%%%%%%%%%%%%%%%%%%%%%%%%%%%%%%%%%%%%%%%%%%%%%%%%%%%%%%%%%
%%%%%%%%%%%%%%%%%%%%%%%%%%%%%%%%%%%%%%%%%%%%%%%%%%%%%%%%%%%%%%%%%%
%%%%%%%%%%%%%%%%%%%%%%%%%%%%%%%%%%%%%%%%%%%%%%%%%%%%%%%%%%%%%%%%%%
%%%%%%%%%%%%%%%%%%%%%%%%%%%%%%%%%%%%%%%%%%%%%%%%%%%%%%%%%%%%%%%%%%
% Bibliography
{
\bibliographystyle{ieee_fullname}
\bibliography{references}
}

\end{document}

%%%%%%%%% TITLE
\title{
    \Large
    Joint Graph-based Depth Refinement and Normal Estimation \\
    -- Supplementary Material -- \\[20pt]
}

%%%%%%%%% AUTHORS
\author{Mattia Rossi\textsuperscript{*}, Mireille El Gheche\textsuperscript{*}, Andreas Kuhn\textsuperscript{\dag}, Pascal Frossard\textsuperscript{*} \\[5pt]
\textsuperscript{*}\'{E}cole Polytechnique F\'{e}d\'{e}rale de Lausanne, \textsuperscript{\dag}Sony Europe B.V.}

\maketitle
\thispagestyle{empty}

\vspace{80pt}
\tableofcontents

%%%%%%%%%%%%%%%%%%%%%%%%%%%%%%%%%%%%%%%%%%%%%%%%%%%%%%%%%%%%%%%%%%
%%%%%%%%%%%%%%%%%%%%%%%%%%%%%%%%%%%%%%%%%%%%%%%%%%%%%%%%%%%%%%%%%%
%%%%%%%%%%%%%%%%%%%%%%%%%%%%%%%%%%%%%%%%%%%%%%%%%%%%%%%%%%%%%%%%%%
%%%%%%%%%%%%%%%%%%%%%%%%%%%%%%%%%%%%%%%%%%%%%%%%%%%%%%%%%%%%%%%%%%
% Inverse depth plane proof
\newpage
\section{Proof of equivalence between Eq.~4 and Eq.~6 of the article}

In Section~3 of our article, we claimed that Eq.~6, which reads as follows,
\begin{equation} \label{eq:inverse_depth_plane}
    d \left( x, y \right) =
    d \left( x_{0}, y_{0} \right) + \langle \left( u(x_{0}, y_{0} \right), \left( x - x_{0}, y - y_{0} \right) \rangle,
\end{equation}
with $u(x_{0}, y_{0}) = (u_{0}^{x}, u_{0}^{y})$ defined as
\begin{equation} \label{eq:inverse_depth_plane_normal}
    u_{0}^{x} = \frac{a_{0}}{\rho_{0} f^{x}}, \quad
    u_{0}^{y} = \frac{b_{0}}{\rho_{0} f^{y}},
\end{equation}
is equivalent to Eq.~4, which reads as follows,
\begin{equation} \label{eq:3d_space_plane}
    \left(
    a_{0} \frac{\left( x - c^{x} \right)}{f^{x}} + b_{0} \frac{\left( y - c^{y} \right)}{f^{y}} + c_{0}
    \right) Z - \rho_{0} = 0,
\end{equation}
with $\rho_{0}$ defined as
\begin{equation} \label{eq:rho}
    \rho_{0} = \left(
    a_{0} \frac{\left( x_{0} - c^{x} \right)}{f^{x}} +
    b_{0} \frac{\left( y_{0} - c^{y} \right)}{f^{y}} +
    c_{0} \right) Z_{0}.
\end{equation}
Here we provide the proof.
\begin{proof}
Let us start by expanding Eq.~\eqref{eq:inverse_depth_plane} as follows:
\begin{align}
    d \left( x, y \right)
    &=
    d \left( x_{0}, y_{0} \right) + \langle \left( u(x_{0}, y_{0} \right), \left( x - x_{0}, y - y_{0} \right) \rangle
    \nonumber \\
    &=
    d \left( x_{0}, y_{0} \right) + \langle \left( u_{0}^{x}, u_{0}^{y} \right), \left( x - x_{0}, y - y_{0} \right) \rangle
    \nonumber \\
    &=
    d \left( x_{0}, y_{0} \right) +
    u_{0}^{x} \left( x - x_{0} \right) +
    u_{0}^{y} \left( y - y_{0} \right)
    \nonumber \\
    &=
    u_{0}^{x} x + u_{0}^{y} y + \left( d \left( x_{0}, y_{0} \right) - u_{0}^{x} x_{0} - u_{0}^{y} y_{0} \right)
    \nonumber \\
    &=
    \left( \frac{a_{0}}{\rho_{0} f^{x}} \right) x + \left( \frac{b_{0}}{\rho_{0} f^{y}} \right) y +
    \left(
    d \left( x_{0}, y_{0} \right) -
    \left( \frac{a_{0}}{\rho_{0} f^{x}} \right) x_{0} - \left( \frac{b_{0}}{\rho_{0} f^{y}} \right) y_{0}
    \right),
    \label{eq:inverse_depth_plane_exapanded}
\end{align}
where the last equality employs Eq.~\eqref{eq:inverse_depth_plane_normal}.
Now, let us express Eq.~\eqref{eq:3d_space_plane} using the \textit{inverse depth} $d(x, y) = 1 / Z(x, y)$:
\begin{align}
    d \left( x, y \right)
    &=
    \frac{1}{\rho_{0}} \left(
    a_{0} \frac{\left( x - c^{x} \right)}{f^{x}} + b_{0} \frac{\left( y - c^{y} \right)}{f^{y}} + c_{0}
    \right) \nonumber \\
    &=
    \left( \frac{a_{0}}{\rho_{0} f^{x}} \right) x +
    \left( \frac{b_{0}}{\rho_{0} f^{y}} \right) y +
    \frac{1}{\rho_{0}} \left( c_{0} - \frac{a_{0} c^{x}}{f^{x}} - \frac{b_{0} c^{y}}{f^{y}} \right). \label{eq:eq:3d_space_plane_expanded}
\end{align}
In order to prove that Eq.~\eqref{eq:inverse_depth_plane} and \eqref{eq:3d_space_plane} are equivalent, it is sufficient to show that the following equality (from Eqs.~\eqref{eq:inverse_depth_plane_exapanded} and \eqref{eq:eq:3d_space_plane_expanded}) holds true:
\begin{equation} \label{eq:key_equality}
    \left(
    d \left( x_{0}, y_{0} \right) -
    \left( \frac{a_{0}}{\rho_{0} f^{x}} \right) x_{0} - \left( \frac{b_{0}}{\rho_{0} f^{y}} \right) y_{0}
    \right) =
    \frac{1}{\rho_{0}} \left( c_{0} - \frac{a_{0} c^{x}}{f^{x}} - \frac{b_{0} c^{y}}{f^{y}} \right).
\end{equation}
We proceed by developing the left part of Eq.~\eqref{eq:key_equality} and use the notation $d_{0} = d(x_{0}, y_{0})$:
\begin{align*}
    \left(
    d_{0} -
    \left( \frac{a_{0}}{\rho_{0} f^{x}} \right) x_{0} - \left( \frac{b_{0}}{\rho_{0} f^{y}} \right) y_{0}
    \right)
    &=
    \frac{f^{x} f^{y} \rho_{0} d_{0} - a_{0} f^{y} x_{0} - b_{0} f^{x} y_{0}}{\rho_{0} f^{x} f^{y}}\\
    &=
    \frac{
    f^{x} f^{y} \left(
    \frac{a_{0} \left( x_{0} - c^{x} \right) }{f^{x} d_{0}} +
    \frac{b_{0} \left( y_{0} - c^{y} \right) }{f^{y} d_{0}} +
    \frac{c_{0}}{d_{0}}
    \right) d_{0} - a_{0} f^{y} x_{0} - b_{0} f^{x} y_{0}
    }{\rho_{0} f^{x} f^{y}}\\
    &=
    \frac{
    a_{0} f^{y} \left( x_{0} - c^{x} \right) +
    b_{0} f^{x} \left( y_{0} - c^{y} \right) +
    f^{x} f^{y} c_{0} - a_{0} f^{y} x_{0} - b_{0} f^{x} y_{0}
    }{\rho_{0} f^{x} f^{y}}\\
    &=
    \frac{
    - a_{0} f^{y} c^{x} - b_{0} f^{x} c^{y} + f^{x} f^{y} c_{0}
    }{\rho_{0} f^{x} f^{y}}\\
    &=
    \frac{1}{\rho_{0}} \left( c_{0} - \frac{a_{0} c^{x}}{f^{x}} - \frac{b_{0} c^{y}}{f^{y}} \right),
\end{align*}
where the second equality comes from Eq.~\eqref{eq:rho}.
\end{proof}

%%%%%%%%%%%%%%%%%%%%%%%%%%%%%%%%%%%%%%%%%%%%%%%%%%%%%%%%%%%%%%%%%%
%%%%%%%%%%%%%%%%%%%%%%%%%%%%%%%%%%%%%%%%%%%%%%%%%%%%%%%%%%%%%%%%%%
%%%%%%%%%%%%%%%%%%%%%%%%%%%%%%%%%%%%%%%%%%%%%%%%%%%%%%%%%%%%%%%%%%
%%%%%%%%%%%%%%%%%%%%%%%%%%%%%%%%%%%%%%%%%%%%%%%%%%%%%%%%%%%%%%%%%%
% Non linear system solution
\newpage
\section{Close form solution of the non linear system in Eq.~7 of the article}

We provide the close form solution of the following non linear system, which permits to recover the normal $n_{0} = (a_{0}, b_{0}, c_{0})$ when the vector $u(x_{0}, y_{0}) = (u_{0}^{x}, u_{0}^{y})$ and $d(x_{0}, y_{0})$ are given:
\begin{subnumcases}{\label{eq:plane_to_space_normal}}
     u_{0}^{x} = \left( \rho_{0} f^{x} \right)^{-1} a_{0} \label{eq:non_linear_sys_1} \\
     u_{0}^{y} = \left( \rho_{0} f^{y} \right)^{-1} b_{0} \label{eq:non_linear_sys_2} \\
     a_{0}^{2} + b_{0}^{2} + c_{0}^{2} = 1. \label{eq:non_linear_sys_3}
\end{subnumcases}
For the case $1$ (i.e., $u_{0}^{x}$, $u_{0}^{y} \neq 0$) we provide both the system solution and its derivation.
For the remaining cases $2$, $3$ and $4$ we provide only the solution, as their derivation follows the one of case $1$.

%%%%%%%%%%%%%%%%%%%%%%%%%%%%%%%%%%%%%%%%%%%%%%%%%%%%%%%%%%%%%%%%%%%%%%%%%%%%%%%%%%%%%%%%%%%%%%%%%%%%
%%%%%%%%%%%%%%%%%%%%%%%%%%%%%%%%%%%%%%%%%%%%%%%%%%%%%%%%%%%%%%%%%%%%%%%%%%%%%%%%%%%%%%%%%%%%%%%%%%%%
\subsection{Case 1: \texorpdfstring{$u_{0}^{x}$}{}, \texorpdfstring{$u_{0}^{y} \neq 0$}{}}

The solution of the system in Eq.~\eqref{eq:plane_to_space_normal} reads as follows:
\begin{equation} \label{eq:non_linear_sys_solution}
    n_{0} = \left[
    \begin{array}{c}
         a_{0} \\
         b_{0} \\
         c_{0}
    \end{array}
    \right] =
    \begin{cases}
    - \left| \gamma \right|
    \left( \left( \alpha + \beta \kappa \right)^{2} + \gamma^{2} \left( 1 + \kappa^{2} \right) \right)^{-1}
    \sign{u_{0}^{x}} \\
    \kappa a_{0} \\
    - \left( \alpha a_{0} + \beta b_{0} \right) \gamma^{-1}.
    \end{cases}
\end{equation}
The used symbols are defined as follows:
\begin{equation} \label{eq:non_linear_sys_solution_notation}
    \begin{aligned}
        \alpha &= u_{0}^{x} f^{y} \left( x_{0} - c^{x} \right) Z_{0} - f^{y}, \\
        \beta &= u_{0}^{x} f^{x} \left(y_{0} - c^{y} \right) Z_{0}, \\
        \gamma &= u_{0}^{x} f^{x} f^{y} Z_{0}, \\
        \delta &= u_{0}^{y} f^{y} \left(x_{0} - c^{x} \right) Z_{0}, \\
        \epsilon &= u_{0}^{y} f^{x} \left( y_{0} - c^{y} \right) Z_{0} - f^{x}, \\
        \phi &= u_{0}^{y} f^{x} f^{y} Z_{0}, \\
        \kappa &= \frac{\alpha \phi - \delta \gamma}{\epsilon \gamma - \beta \phi} =
        \left( u_{0}^{x} f^{x} \right)^{-1} u_{0}^{y} f^{y}.
    \end{aligned}
\end{equation}
Before we proceed to the proof, a remark is needed.
\begin{remark}
In the article, the 3D plane ${\mathscr{P}}$ is identified by the pair $(P_{0} = (X_{0}, Y_{0}, Z_{0}), n_{0})$ with $P_{0} \in \mathscr{P}$.
However, $\mathscr{P}$ can be identified by the pair $(P_{0}, - n_{0})$ as well.
If the plane represents a physical surface, e.g., a wall, then the two normals $n_{0}$ and $- n_{0}$ can be thought as representing the two sides of the plane.
It is reasonable to represent the plane with the normal $n_{0}$ associated to the side of the plane observed by the camera.
This is equivalent to require that the angle between $n_{0}$ and the vector from the point $P_{0}$ to the pinhole camera origin is acute.
Formally, this translates into the following constraint:
\begin{equation} \label{eq:visibility_constraint}
    \underbrace{
        \langle n_{0}, \left( X_{0}, Y_{0}, Z_{0} \right) \rangle
    }_{\rho_{0}} < 0.
\end{equation}
\end{remark}
\begin{proof}
We start by replacing $\rho_{0}$, defined in Eq.~\eqref{eq:rho} in image coordinates, in the system of Eq.~\eqref{eq:plane_to_space_normal}:
\begin{equation} \label{eq:normal_sys_expanded}
    \begin{cases}
        \left( u_{0}^{x} f^{y} \left( x_{0} - c^{x} \right) Z_{0} - f^{y} \right) a_{0} +
        \left( u_{0}^{x} f^{x} \left( y_{0} - c^{y} \right) Z_{0} \right) b_{0} +
        \left( u_{0}^{x} f^{x} f^{y} Z_{0} \right) c_{0} = 0 \\
        \left( u_{0}^{y} f^{y} \left( x_{0} - c^{x} \right) Z_{0} \right) a_{0} +
        \left( u_{0}^{y} f^{x} \left( y_{0} - c^{y} \right) Z_{0} - f^{x} \right) b_{0} +
        \left( u_{0}^{y} f^{x} f^{y} Z_{0} \right) c_{0} = 0 \\
        a_{0}^{2} + b_{0}^{2} + c_{0}^{2} = 1.
    \end{cases}
\end{equation}
Using the notation in Eq.~\eqref{eq:non_linear_sys_solution_notation}, we can rewrite the system in Eq.~\eqref{eq:normal_sys_expanded} as follows:
\begin{subnumcases}{}
    \alpha a_{0} + \beta b_{0} + \gamma c_{0} = 0 \label{eq:non_linear_sys_1_greek} \\
    \delta a_{0} + \epsilon b_{0} + \phi c_{0} = 0 \label{eq:non_linear_sys_2_greek} \\
    a_{0}^{2} + b_{0}^{2} + c_{0}^{2} = 1. \label{eq:non_linear_sys_3_greek}
\end{subnumcases}
Let us isolate $c_{0}$ in Eqs.~\eqref{eq:non_linear_sys_1_greek} and \eqref{eq:non_linear_sys_2_greek}:
\begin{subequations}
    \begin{align}
        c_{0} = - \frac{\alpha a_{0} + \beta b_{0}}{\gamma} \label{eq:c_1} \\
        c_{0} = - \frac{\alpha a_{0} - \epsilon b_{0}}{\phi}. \label{eq:c_2}
    \end{align}
\end{subequations}
We observe that the divisions by $\gamma$ and $\sigma$ are legit, as $\gamma$, $\sigma \neq 0$ holds true.
In fact, in the definitions of $\gamma$ and $\sigma$ in Eqs.~\eqref{eq:non_linear_sys_solution_notation}, the quantities $f^{x}$, $f^{y}$, $Z_{0}$ are positive by definition and $u_{0}^{x}$, $u_{0}^{y} \neq 0$ by assumption.
Summing the Eqs.~\eqref{eq:c_1} and \eqref{eq:c_2} side-wise, and isolating $b_{0}$, leads to the following expression for $b_{0}$:
\begin{equation} \label{eq:b_1}
    b_{0} = \underbrace{\left( \frac{\alpha \phi - \delta \gamma}{\epsilon \gamma - \beta \phi} \right)}_{\kappa} a_{0}.
\end{equation}
From the definitions in Eq.~\eqref{eq:non_linear_sys_solution_notation} we have that $\epsilon \gamma - \beta \phi = - u_{0}^{x} \left( f^{x} \right) ^{2} f^{y} Z_{0} \neq 0$ holds true, hence division is legit.
We replace Eq.~\eqref{eq:b_1} in Eq.~\eqref{eq:non_linear_sys_3_greek} and get the following expression for $c_{0}$:
\begin{equation}
    c_{0} = \pm \sqrt{1 - \left( 1 + \kappa^{2} \right) a_{0}^{2}} \label{eq:c_3}.
\end{equation}
Now, we replace the Eqs.~\eqref{eq:b_1} and \eqref{eq:c_3} in Eq.~\eqref{eq:non_linear_sys_1_greek} and solve for $a_{0}$:
\begin{gather*}
    \alpha a_{0} + \beta \kappa a_{0} + \gamma \sign{c_{0}} \sqrt{1 - \left( 1 + \kappa^{2} \right) a_{0}^{2}} = 0 \nonumber \\
    \left( \alpha + \beta \kappa \right) a_{0} = - \gamma \sign{c_{0}} \sqrt{1 - \left( 1 + \kappa^{2} \right) a_{0}^{2}} \nonumber \\
    \left(\alpha + \beta \kappa \right)^{2} a_{0}^{2} = \gamma^{2} \left( 1 - \left( 1 + \kappa^{2} \right) a_{0}^{2} \right) \nonumber \\
    \left(\alpha + \beta \kappa \right)^{2} a_{0}^{2} + \gamma^{2} \left( 1 + \kappa^{2} \right) a_{0}^{2} = \gamma^{2} \nonumber \\
    a_{0}^{2} \left( \left(\alpha + \beta \kappa \right)^{2} + \gamma^{2} \left( 1 + \kappa^{2} \right) \right) = \gamma^{2} \nonumber \\
    a_{0}^{2} = \frac{\gamma^{2}}{\left(\alpha + \beta \kappa \right)^{2} + \gamma^{2} \left( 1 + \kappa^{2} \right)} \nonumber \\
    a_{0} = \pm \sqrt{\frac{\gamma^{2}}{\left(\alpha + \beta \kappa \right)^{2} + \gamma^{2} \left( 1 + \kappa^{2} \right)}} \nonumber \\
    a_{0} = \pm \frac{\left| \gamma \right|}{\sqrt{\left(\alpha + \beta \kappa \right)^{2} + \gamma^{2} \left( 1 + \kappa^{2} \right)}}.
\end{gather*}
From Eq.~\eqref{eq:non_linear_sys_1} we know that $\sign{a_{0}} = \sign{\rho_{0} f^{x} u_{0}^{x}} = - \sign{u_{0}^{x}}$ holds true, as $f^{x}$ is positive by definition and $\rho_{0} < 0$ according to Eq.~\eqref{eq:visibility_constraint}.
We thus have the following expression for the component $a_{0}$ of the normal:
\begin{equation} \label{eq:a_final}
    a_{0} =
    - \sign{u_{0}^{x}}
    \frac{\left| \gamma \right|}
    {\sqrt{\left(\alpha + \beta \kappa \right)^{2} + \gamma^{2} \left( 1 + \kappa^{2} \right)}}.
\end{equation}
Replacing Eq.~\eqref{eq:a_final} in the Eqs.~\eqref{eq:b_1} and \eqref{eq:non_linear_sys_1_greek} provides the expressions for the components $b_{0}$ and $c_{0}$ of the normal, respectively:
\begin{align*}
    b_{0} &= \kappa a_{0} \\
    c_{0} &= - \frac{\alpha a_{0} + \beta b_{0}}{\gamma}.
\end{align*}
\end{proof}

%%%%%%%%%%%%%%%%%%%%%%%%%%%%%%%%%%%%%%%%%%%%%%%%%%%%%%%%%%%%%%%%%%%%%%%%%%%%%%%%%%%%%%%%%%%%%%%%%%%%
%%%%%%%%%%%%%%%%%%%%%%%%%%%%%%%%%%%%%%%%%%%%%%%%%%%%%%%%%%%%%%%%%%%%%%%%%%%%%%%%%%%%%%%%%%%%%%%%%%%%
\subsection{Case 2: \texorpdfstring{$u_{0}^{x} \neq 0$}{}, \texorpdfstring{$u_{0}^{y} = 0$}{}}

The solution of the system in Eq.~\eqref{eq:plane_to_space_normal} reads as follows:
\begin{equation}
    n_{0} = \left[
    \begin{array}{c}
         a_{0} \\
         b_{0} \\
         c_{0}
    \end{array}
    \right] =
    \begin{cases}
    - \left| \gamma \right|
    \left( \sqrt{\alpha^{2} + \gamma^{2}} \right)^{-1}
    \sign{u_{0}^{x}} \\
    0 \\
    - \left( \alpha a_{0} \right) \gamma^{-1}.
    \end{cases}
\end{equation}

%%%%%%%%%%%%%%%%%%%%%%%%%%%%%%%%%%%%%%%%%%%%%%%%%%%%%%%%%%%%%%%%%%%%%%%%%%%%%%%%%%%%%%%%%%%%%%%%%%%%
%%%%%%%%%%%%%%%%%%%%%%%%%%%%%%%%%%%%%%%%%%%%%%%%%%%%%%%%%%%%%%%%%%%%%%%%%%%%%%%%%%%%%%%%%%%%%%%%%%%%
\subsection{Case 3: \texorpdfstring{$u_{0}^{x} = 0$}{}, \texorpdfstring{$u_{0}^{y} \neq 0$}{}}

The solution of the system in Eq.~\eqref{eq:plane_to_space_normal} reads as follows:
\begin{equation}
    n_{0} = \left[
    \begin{array}{c}
         a_{0} \\
         b_{0} \\
         c_{0}
    \end{array}
    \right] =
    \begin{cases}
    0 \\
    - \left| \phi \right|
    \left( \sqrt{\epsilon^{2} + \phi^{2}} \right)^{-1}
    \sign{u_{0}^{y}} \\
    - \left( \epsilon b_{0} \right) \phi^{-1}.
    \end{cases}
\end{equation}

%%%%%%%%%%%%%%%%%%%%%%%%%%%%%%%%%%%%%%%%%%%%%%%%%%%%%%%%%%%%%%%%%%%%%%%%%%%%%%%%%%%%%%%%%%%%%%%%%%%%
%%%%%%%%%%%%%%%%%%%%%%%%%%%%%%%%%%%%%%%%%%%%%%%%%%%%%%%%%%%%%%%%%%%%%%%%%%%%%%%%%%%%%%%%%%%%%%%%%%%%
\subsection{Case 4: \texorpdfstring{$u_{0}^{x}$}{}, \texorpdfstring{$u_{0}^{y} = 0$}{}}

The solution of the system in Eq.~\eqref{eq:plane_to_space_normal} reads as follows:
\begin{equation}
    n_{0} = \left[
    \begin{array}{c}
         a_{0} \\
         b_{0} \\
         c_{0}
    \end{array}
    \right] =
    \begin{cases}
    \phantom{+} 0 \\
    \phantom{+} 0 \\
    - 1.
    \end{cases}
\end{equation}

%%%%%%%%%%%%%%%%%%%%%%%%%%%%%%%%%%%%%%%%%%%%%%%%%%%%%%%%%%%%%%%%%%
%%%%%%%%%%%%%%%%%%%%%%%%%%%%%%%%%%%%%%%%%%%%%%%%%%%%%%%%%%%%%%%%%%
%%%%%%%%%%%%%%%%%%%%%%%%%%%%%%%%%%%%%%%%%%%%%%%%%%%%%%%%%%%%%%%%%%
%%%%%%%%%%%%%%%%%%%%%%%%%%%%%%%%%%%%%%%%%%%%%%%%%%%%%%%%%%%%%%%%%%
% Disparity up-sampling
\newpage
\section{Proof that the inverse depth map up-sampling requires scaling}

In the article we claim that the up-sampling of a planar inverse depth $d^{\ell}$ with slope $u^{\ell}$ leads to a new planar inverse depth $d^{\ell - 1}$ with slope ${u^{\ell - 1} = r^{-1} u^{\ell}}$.
Here we provide the proof.
\begin{proof}
Let us recall the equation of a planar inverse depth map at scale $\ell$:
\begin{equation}
    d^{\ell} \left( x, y \right) = d^{\ell} \left( x_{0}, y_{0} \right) + \langle u^{\ell}, \left( x - x_{0}, y - y_{0} \right) \rangle,
\end{equation}
where we remove the dependencies of $u^{\ell}$ from $(x_{0}, y_{0})$, as $u^{\ell}(x_{0}, y_{0})$ is constant for a planar inverse depth map.
In the multi-scale approach, up-sampling $d^{\ell}$ by a factor $r$ is equivalent to decrease the pixel size by a factor $r$ along both the pixel dimensions, as the camera sensor dimensions do not change.
It follows that up-sampling $d^{\ell}$ by a factor $r$ is equivalent to re-sampling it with a step $r^{-1}$:
\begin{equation*}
    \begin{aligned}
        d^{\ell - 1} \left( x, y \right) = d^{\ell} \left( \frac{x}{r}, \frac{y}{r} \right)
        &= d^{\ell} \left( x_{0}, y_{0} \right) + \langle u^{\ell}, \left( \frac{x}{r} - x_{0}, \frac{y}{r} - y_{0} \right) \rangle \\
        &= d^{\ell} \left( x_{0}, y_{0} \right) + \langle \frac{u^{\ell}}{r}, \left( x, y \right) \rangle - \langle u^{\ell}, \left( x_{0}, y_{0} \right) \rangle \\
        &= d^{\ell - 1} \left( r x_{0}, r y_{0} \right) + \langle \frac{u^{\ell}}{r} \left( x - r x_{0}, y - r y_{0} \right) \rangle \\
        &= d^{\ell - 1} \left( \hat{x}_{0}, \hat{y}_{0} \right) + \langle  \underbrace{r^{-1} u^{\ell}}_{u^{\ell -1}} \left( x - \hat{x}_{0}, y - \hat{y}_{0} \right) \rangle,
    \end{aligned}
\end{equation*}
where in the last equality we defined $\hat{x}_{0} = r x_{0}$ and $\hat{y}_{0} = r y_{0}$.
\end{proof}

%%%%%%%%%%%%%%%%%%%%%%%%%%%%%%%%%%%%%%%%%%%%%%%%%%%%%%%%%%%%%%%%%%
%%%%%%%%%%%%%%%%%%%%%%%%%%%%%%%%%%%%%%%%%%%%%%%%%%%%%%%%%%%%%%%%%%
%%%%%%%%%%%%%%%%%%%%%%%%%%%%%%%%%%%%%%%%%%%%%%%%%%%%%%%%%%%%%%%%%%
%%%%%%%%%%%%%%%%%%%%%%%%%%%%%%%%%%%%%%%%%%%%%%%%%%%%%%%%%%%%%%%%%%
% Regularization analysis
\newpage
\section{Comparison between the proposed regularization and NLTGV}

The regularization proposed in the article can be rewritten, more explicitly, as follows:
\begin{align}
    g \left( d, u \right) &=
    \sum_{i}
    \left \Vert \underbrace{
        \begin{pmatrix}
            w_{i j_{1}} \left( d_{j_{1}} - d_{i} - \langle u_{i}, j_{1} - i \rangle \right) \\
            w_{i j_{2}} \left( d_{j_{2}} - d_{i} - \langle u_{i}, j_{2} - i \rangle \right) \\
            \vdots \\
            w_{i j_{K_{i}}} \left( d_{j_{K_{i}}} - d_{i} - \langle u_{i}, j_{K_{i}} - i \rangle \right)
        \end{pmatrix}}_{v_{i}}
    \right \Vert_{2} \label{eq:prior_planes} \\
    &+
    \alpha \sum_{i}
    \left \Vert
        \begin{pmatrix}
            w_{i j_{1}} \Vert u_{j_{1}} - u_{i} \Vert_{2} \\
            w_{i j_{2}} \Vert u_{j_{2}} - u_{i} \Vert_{2}\\
            \vdots \\
            w_{i j_{K_{i}}} \Vert u_{j_{K_{i}}} - u_{i} \Vert_{2}
        \end{pmatrix}
    \right \Vert_{1}, \label{eq:prior_normals}
\end{align}
where $j_{1}, j_{2}, \ldots, j_{K_{i}}$ are the $K_{i} \in \mathbb{N}$ pixels directly connected to $i$ in the graph, i.e., pixel $i$ neighbourhood, while $w_{i j} \in \mathbb{R}_{>0}$ with $j \in  \{ j_{1}, j_{2}, \ldots, j_{K_{i}} \}$ is the weight associated to the edge between the pixel $i$ and the neighboring pixel $j$.
Our regularization in Eqs.~\eqref{eq:prior_planes} and \eqref{eq:prior_normals} resembles \textit{Non Local Total Generalized Variation (NLTGV)} \cite{ranftl_non_local_2014}, which reads as follows:
\begin{align}
    g_{NLTGV} \left( d, u \right) &=
    \sum_{i}
    \left \Vert \underbrace{
        \begin{pmatrix}
            w_{i j_{1}} \left( d_{j_{1}} - d_{i} - \langle u_{i}, j_{1} - i \rangle \right) \\
            w_{i j_{2}} \left( d_{j_{2}} - d_{i} - \langle u_{i}, j_{2} - i \rangle \right) \\
            \vdots \\
            w_{i j_{K_{i}}} \left( d_{j_{K_{i}}} - d_{i} - \langle u_{i}, j_{K_{i}} - i \rangle \right)
        \end{pmatrix}}_{v_{i}}
    \right \Vert_{1} \label{eq:prior_planes_nltgv} \\
    &+
    \alpha \sum_{i}
    \left \Vert
        \begin{pmatrix}
            w_{i j_{1}} \left(
                \left| u^{x}_{j_{1}} - u^{x}_{i} \right| + \left| u^{y}_{j_{1}} - u^{y}_{i} \right| \right) \\
            w_{i j_{2}} \left( \left| u^{x}_{j_{2}} - u^{x}_{i} \right| + \left| u^{y}_{j_{2}} - u^{y}_{i} \right| \right) \\
            \vdots \\
            w_{i j_{K_{i}}} \left( \left| u^{x}_{j_{K_{i}}} - u^{x}_{i} \right| + \left| u^{y}_{j_{K_{i}}} - u^{y}_{i} \right| \right) \\
        \end{pmatrix}
    \right \Vert_{1} \label{eq:prior_normals_nltgv}
\end{align}

Let us start by comparing the first term of NLTGV and of our regularizer, in Eqs.~\eqref{eq:prior_planes_nltgv} and \eqref{eq:prior_planes}, respectively.
The two terms differ in the norm used to aggregate the entries of the vector $v_{i}$: the NLTGV regularization employs an $\ell_1$--norm, while ours employs an $\ell_2$.
The use of the $\ell_1$--norm in the NLTGV term in Eq.~\eqref{eq:prior_planes_nltgv} permits to rewrite it as follows:
\begin{align} \label{eq:prior_planes_nltgv_2}
    \sum_{i} \sum_{k = 1}^{K_{i}}
        \left| 
        w_{i j_{k}} \left( d_{j_{k}} - d_{i} - \langle u_{i}, j_{k} - i \rangle \right) \right|,
\end{align}
which can be still interpreted as an $\ell_1$--norm, in particular applied to the vector containing all the possible entries $w_{i j_{k}}(d_{j_{k}} - d_{i} - \langle u_{i}, j_{k} - i \rangle$.
The $\ell_1$--norm is known to promote sparse vectors, therefore the minimization of the function in Eq.~\eqref{eq:prior_planes_nltgv_2} would try to zero the terms $w_{i j_{k}}(d_{j_{k}} - d_{i} - \langle u_{i}, j_{k} - i \rangle$ independently.
Equivalently, the fulfillment of the constraints
\begin{equation} \label{eq:inverse_depth_plane_constraint}
    d_{j} = d_{i} + \langle u_{i}, j - i \rangle \qquad\qquad \forall j \in \{j_{1}, \ldots, j_{K_{i}}\}
\end{equation}
would be treated separately for each pair $(i, j)$, which could potentially lead to a misfitted plane.
On the other hand, the first term of our regularization, in Eq.~\eqref{eq:prior_planes}, can be rewritten as follows:
\begin{align} \label{eq:prior_planes_2}
    \sum_{i} \Vert v_{i} \Vert_{2},
\end{align}
which can be interpreted as an $\ell_1$--norm too, but applied to the vector $[\| v_{1}^{\top} \|_{2}, \ldots, \| v_{i}^{\top} \|_{2}, \ldots, \| v_{N^{2}}^{\top} \|_{2}]$ with $v_{i} \in \mathbb{R} ^{K_{i}}$, and it is referred to as the mixed norm $\ell_{1,2}$ \cite{bach_optimization_2012} \cite{rossi_nonsmooth_2018}.
The minimization of the function in Eq.~\eqref{eq:prior_planes_2} would try to zero the entries $v_{i}$ independently, but zeroing an entry $v_{i}$ translates into fulfilling all the constraints in Eq.~\eqref{eq:inverse_depth_plane_constraint} at once, which permits to fit a plane considering all the $K_{i}$ neighboring pixels.

The second term of NLTGV and our regularization, in Eqs.~\eqref{eq:prior_normals_nltgv} and \eqref{eq:prior_normals}, respectively, are both examples of \textit{Non Local Total Variation NLTV} \cite{gilboa_nonlocal_2009}, whose aim is to promote, directly, a piece-wise constant field $u$ and therefore to promote, indirectly, a piece-wise planar inverse depth map $d$.
Differently from the term of NLTGV in Eq.~\eqref{eq:prior_normals_nltgv}, our term in Eq.~\eqref{eq:prior_normals} aggregates the two spatial components of $u_{i}$ using an $\ell_{2}$--norm, which leads the overall term in Eq.~\eqref{eq:prior_normals} to be a mixed $\ell_{1, 2}$--norm as well.
However, differently from the first term of our regularization, where the $\ell_{2}$--norm aggregation is applied to $i$'s neighborhood, here it is limited to the spatial components of $u_{i}$.
The possible benefits of extending the $\ell_{2}$--norm aggregation at the neighborhood level is left to a future investigation.

%%%%%%%%%%%%%%%%%%%%%%%%%%%%%%%%%%%%%%%%%%%%%%%%%%%%%%%%%%%%%%%%%%
%%%%%%%%%%%%%%%%%%%%%%%%%%%%%%%%%%%%%%%%%%%%%%%%%%%%%%%%%%%%%%%%%%
%%%%%%%%%%%%%%%%%%%%%%%%%%%%%%%%%%%%%%%%%%%%%%%%%%%%%%%%%%%%%%%%%%
%%%%%%%%%%%%%%%%%%%%%%%%%%%%%%%%%%%%%%%%%%%%%%%%%%%%%%%%%%%%%%%%%%
% Bibliography
\newpage
{
\bibliographystyle{ieee_fullname}
\bibliography{references.bib}
}